\documentclass[11pt]{article}

\usepackage[final]{acl}

\usepackage{times}
\usepackage{latexsym}

\usepackage[T1]{fontenc}

\usepackage[utf8]{inputenc}

\usepackage{microtype}

\usepackage{inconsolata}

\usepackage{graphicx}
\usepackage{amsmath}
\usepackage{amssymb}
\usepackage{subcaption}
\usepackage{algorithm}
\usepackage{algpseudocode}
\usepackage[many]{tcolorbox}
\usepackage{listings}
\usepackage{enumitem}
\DeclareMathOperator*{\argmax}{arg\,max}

\title{Retrieval Heads are Dynamic}

\author{
    \textbf{Yuping Lin\textsuperscript{1}}\thanks{Work done during internship at Tongyi Lab, Alibaba Group.},
    \textbf{Zitao Li\textsuperscript{2}}\thanks{Work done during at Tongyi Lab, Alibaba Group.},
    \textbf{Yue Xing\textsuperscript{1}},
    \textbf{Pengfei He\textsuperscript{1}},
    \textbf{Yingqian Cui\textsuperscript{1}}, \\
    \textbf{Yaliang Li\textsuperscript{3}},
    \textbf{Bolin Ding\textsuperscript{3}},
    \textbf{Jingren Zhou\textsuperscript{3}},
    \textbf{Jiliang Tang\textsuperscript{1}},
    \\
    \\
    \textsuperscript{1}Michigan State University,
    \textsuperscript{2}Zoom Communications,
    \textsuperscript{3}Tongyi Lab, Alibaba Group,
    \\
    \\
    \texttt{\{linyupin, xingyue1, hepengf1, cuiyingq, tangjili\}@msu.edu,} \\
    \texttt{zitao.li@zoom.us,} \\
    \texttt{\{yaliang.li, bolin.ding, jingren.zhou\}@alibaba-inc.com}
}

\begin{document}
\maketitle
\begin{abstract}

Recent studies have identified ``retrieval heads'' in Large Language Models (LLMs) responsible for extracting information from input contexts. However, prior works largely rely on static statistics aggregated across datasets, identifying heads that perform retrieval on average. This perspective overlooks the fine-grained temporal dynamics of autoregressive generation.
In this paper, we investigate retrieval heads from a dynamic perspective. Through extensive analysis, we establish three core claims: (1) Dynamism: Retrieval heads vary dynamically across timesteps; (2) Irreplaceability: Dynamic retrieval heads are specific at each timestep and cannot be effectively replaced by static retrieval heads; and (3) Correlation: The model's hidden state encodes a predictive signal for future retrieval head patterns, indicating an internal planning mechanism. We validate these findings on the Needle-in-a-Haystack task and a multi-hop QA task, and quantify the differences on the utility of dynamic and static retrieval heads in a Dynamic Retrieval-Augmented Generation framework. Our study provides new insights into the internal mechanisms of LLMs.

\end{abstract}
\section{Introduction}\label{sec:intro}


\begin{figure}[h!]
    \centering
    \includegraphics[width=0.95\linewidth]{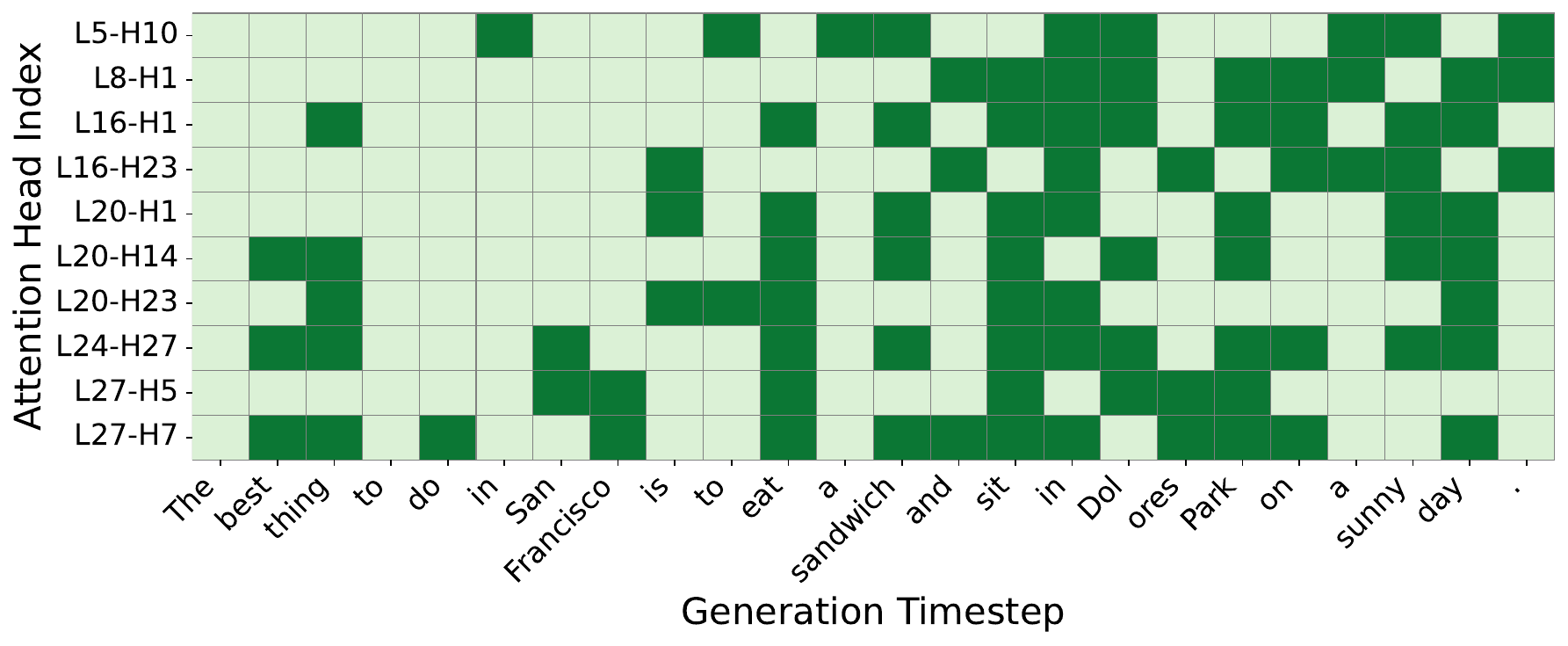}
    \caption{\textbf{Dynamism of Retrieval Heads.} The retrieval scores of individual attention heads fluctuate across the generation process. \textbf{Dark color} indicates heads having a retrieval score of 1, as defined by Equation~\eqref{eq:retrieval_score_niah}. The x-axis denotes the generation step, labeled by the token generated at that step. The y-axis shows the 10 most variated retrieval heads, selected based on their retrieval score variance over the entire generation process. L$x$-H$y$ denotes the $y$-th head (starting from 0) on the layer $x$ (starting from 0).}
    \label{fig:retrieval_dynamics}\vspace{-0.15in}
\end{figure}

Recently, there is a growing interest in Large Language Models (LLMs)~\citep{radford2019language,brown2020language,vaswani2017attention,chowdhery2023palm,hoffmann2022training,touvron2023llama} to understand how they process context, particularly focusing on their ability to extract key information from the input. 
Prior work has shown that although LLMs demonstrate strong in-context learning abilities~\citep{garg2022can, xie2021explanation},
their ability to utilize long contexts often needs improvement~\citep{liu2024lost, press2023measuring}.
A line of mechanistic interpretability works suggests that attention heads exhibit functional specialization~\citep{voita2019analyzing,michel2019sixteen,
elhage2021mathematical}:
For example, a pioneer work~\citep{wu2024retrieval} analyzes the model from an attention-head perspective, identifying a specific set of heads termed ``retrieval heads'' that are responsible for the copy-paste behavior of LLMs from the given inputs. 
Recent studies~\citep{zhang2025query, fu2024not} provide further evidence of the existence of retrieval heads, even in tasks requiring complex reasoning.

While these works provide valuable insights on the mechanism of LLMs, they have been mainly constrained to a \textbf{fixed} subset of attention heads.
For example, \citet{wu2024retrieval} aggregated attention patterns across datasets to find the heads which frequently perform copy-paste operations, identifying a fixed set of heads for each model that perform retrieval on average. 
However, treating retrieval heads as a fixed set assumes that this average behavior is a heuristic approximation for the model's real-time operation. Given the autoregressive nature of LLMs, it is natural to question whether the retrieval heads should instead be a \textbf{dynamic} set conditioned on the given context. Relying on static definitions risks oversimplifying the model's real-time behavior, as heads that are statistically dominant may not be active at every critical timestep, while ``less significant'' heads might play irreplaceable roles in specific contexts.
As illustrated in Figure~\ref{fig:retrieval_dynamics}, the set of heads acting as retrieval heads actually fluctuates significantly across token generation steps. 
This observation challenges the completeness of static definitions, raising the following fundamental questions:
\noindent\textit{1) How does the set of retrieval heads evolve during generation?}
\textit{2) Are these dynamic heads functionally interchangeable with static ones?}
\textit{3) Is this dynamism predictable given a model and a context?}

To answer these questions, we present the first systematic study of retrieval heads from a \textbf{dynamic} perspective. 



Our key contributions are as follows: 

\begin{enumerate}[leftmargin=*]
    \item We demonstrate that the retrieval heads are highly \textbf{dynamic} that statistical methods fail to capture.
    (Claim 1 in Section \ref{sec:analysis_dynamism})
    \item Given the dynamic nature of the retrieval head pattern, we show that the specific retrieval heads at a given generation step are \textbf{not replaceable}, and ablating them causes severe performance degradation. (Claim 2 in Section \ref{sec:analysis_irreplaceable})
    \item We reveal that the model's final hidden state exhibits a strong \textbf{correlation} with future retrieval head patterns, revealing a \textbf{predictive mechanism} within LLMs. (Claim 3 in Section \ref{sec:analysis_predictable})
    \item While the above claims are based on a Needle-in-a-Haystack task in Section \ref{sec:analysis}, we validate them in a question-answering task where reasoning efforts are needed. (Section \ref{sec:hotpotqa})
    \item We use the dynamic and static retrieval heads in a Dynamic RAG scenario to compare their practical utility, demonstrating that dynamically selecting heads based on the current generative state significantly improves retrieval accuracy and downstream performance compared to static retrieval heads. (Section \ref{sec:experiment})
\end{enumerate}



We hope these findings will serve as a foundation for future research in model interpretability and the development of more precise, state-aware intervention techniques.

\vspace{-0.05in}
\section{Related Works}


\paragraph{Mechanistic Interpretability} Understanding the internal mechanisms of Transformer-based models~\citep{vaswani2017attention} has been a focal point of recent research. 
Mechanistic interpretability has emerged as a principled approach to understanding neural networks beyond input-output behavior~\citep{olah2020zoom,elhage2021mathematical}. 
Early work by \citet{olsson2022context} identified Induction Heads, a specialized circuit responsible for in-context learning by copying previous tokens that follow similar patterns. 
Subsequent work further 
investigated induction-like circuits~\citep{elhage2022toy,
wang2022interpretability}. 
This laid the theoretical groundwork for understanding how attention heads perform ``copy-paste'' operations. In the context of long-sequence modeling, \citet{xiao2023efficient} discovered Attention Sinks, revealing that models dedicate massive attention to initial tokens (e.g., BOS) to maintain numerical stability.
Related analyses have also examined positional and numerical artifacts in attention mechanisms~\citep{press2021train,su2024roformer}.
Furthermore, studies on the ``Lost in the Middle'' phenomenon~\citep{liu2024lost} have highlighted the non-uniform capability of models to access information across long contexts.
These studies primarily focus on static circuit structures or attention biases, and do not fully explain how the model dynamically modulates its attention allocation step-by-step to perform precise retrieval during the autoregressive generation.

\paragraph{Retrieval Heads} 
Information retrieval within LLMs has been studied both implicitly through attention mechanisms and explicitly through retrieval-augmented generation (RAG) frameworks~\citep{lewis2020retrieval,borgeaud2022improving}.
Building on the concept of functional specialization, recent studies have isolated specific attention heads responsible for information retrieval. \citet{wu2024retrieval} pioneered this direction by identifying Retrieval Heads via the Needle-in-a-Haystack (NIAH) test, characterizing them as a sparse, intrinsic subset of heads that perform copy-paste operations from long contexts. Addressing the limitations of synthetic benchmarks, \citet{zhang2025query} proposed QRHead, which refines head detection using query-aware attention scores on realistic tasks to improve downstream retrieval and re-ranking performance. Similarly, \citet{fu2024not} introduced HeadKV, a method that leverages retrieval and reasoning importance scores to perform head-level KV cache compression, significantly outperforming layer-level methods like SnapKV~\citep{li2024snapkv}, H2O~\citep{zhang2023h2o}, and PyramidKV~\citep{cai2024pyramidkv}. A common limitation across these works is their reliance on a \textit{static perspective}. However, this approach overlooks the \textit{temporal dynamism} of the generation process.

\vspace{-0.05in}
\section{Analysis of Dynamic Retrieval Heads}\label{sec:analysis}

\subsection{Setup}

This section focuses on the traditional \textbf{copy-paste retrieval head} as in~\citet{wu2024retrieval}. This type of retrieval head considers the exact copy-paste of the input tokens to the next generated token. 

To better trace and analyze the exact copy-paste behavior, following~\citet{wu2024retrieval}, we consider the Needle-in-a-Haystack (NIAH) task~\citep{kamradt2023niah}, which evaluates a model's ability to precisely retrieve the specific piece of information (the ``needle'') embedded at a random location within a long, distracting document (the ``haystack'').

\paragraph{Definition of {Retrieval Head}} Following~\citet{wu2024retrieval}, to define the copy-paste retrieval head, an attention head is considered to be performing a retrieval operation if and only if two conditions are satisfied: (1) at the current inference step, the generated token is identical to the token receiving the highest attention weight from that head, and (2) the token with the highest attention weight lies within the ``needle'' context, i.e., it is a \textit{needle token}. 
When these conditions are satisfied, its retrieval score is set to be 1.\footnote{The original work~\citep{wu2024retrieval} normalizes this score by the length of the needle. We omit this normalization, assigning a binary score of 1 and 0, because our analysis is conducted at the token level, in contrast to the sample-level analysis of the original work.} 



Formally, let $i^* = \argmax_i (\mathbf{a}^{h,t}_i)$ be the index of the token that receives the maximum attention from head $h$ at timestep $t$, where $\mathbf{a}^{h,t}$ is the vector of attention scores from the final token of the input $x^t$ (as query) to all tokens in $x^t$ (as keys) for head $h$, i.e., $\mathbf{a}^{h,t} = \mathrm{AttnScore}^{h}[t, :]$.
The retrieval score of head $h$ on input $x^t$ is then defined as:
\vspace{-0.05in}
\begin{equation}\label{eq:retrieval_score_niah}
    S_{\mathrm{copy-paste}}(x^t, h) = \mathbf{1} \left[ i^* \in I_{\text{needle}} \land x^t_{i^*} = \hat{y} \right]
\end{equation}
\noindent where $I_{\text{needle}}$ is the set of indices for tokens within the needle, and $\hat y$ is the token predicted to be generated at the current timestep. 

\paragraph{Overview of Claims}
Given the above task description and definition of retrieval heads, we present our central claims:

\begin{itemize}[leftmargin=*]\vspace{-0.1in}
    \item \textbf{Claim 1: Dynamism.} The patterns of retrieval heads are dynamic throughout the autoregressive generation process. \vspace{-0.05in}
    \item \textbf{Claim 2: Irreplaceability.} The retrieval functionality of the specific retrieval heads at a given timestep cannot be replaced by other heads. If these heads are disabled, the model will suffer from performance degradation.\vspace{-0.05in}
    \item \textbf{Claim 3: Correlation.} A strong correlation exists between the model's hidden state and the patterns of retrieval heads in the future.
\end{itemize}


\subsection{Retrieval Heads are Dynamic}\label{sec:analysis_dynamism}

Different from existing literature \citep{wu2024retrieval, zhang2025query, fu2024not} where a large corpus of samples are collected to identify a set of statistically significant retrieval heads (i.e., \textbf{static retrieval heads}), we argue the existence of unique patterns of retrieval heads that emerge at individual timesteps during the generation process (i.e., \textbf{dynamic retrieval heads}). Specifically, the retrieval score of attention heads fluctuates across timesteps. Therefore, we hypothesize that the dynamic retrieval heads at a particular timestep do not always align with the static retrieval heads.

To verify our hypothesis, Figure~\ref{fig:retrieval_dynamics} in Section \ref{sec:intro} plots the retrieval scores calculated as per Equation~\eqref{eq:retrieval_score_niah} for some attention heads over the course of an autoregressive generation process for a given sample. 
The plot clearly demonstrates that the retrieval scores for individual heads fluctuate significantly across timesteps, confirming the dynamic nature of retrieval heads.

Furthermore, to rigorously quantify this dynamism, we conducted a statistical analysis. The results for all models are summarized in Table~\ref{tab:dynamism_stats}.

\begin{table}[h]
    \centering
    \resizebox{1.0\linewidth}{!}{
    \begin{tabular}{lccccc}
        \hline
        Model & Mean $\pm$ Std & Unique / Total & Jaccard w/ Static & Adj. Jaccard & Entropy \\
        \hline
        llama3.1-8b & 12.97 $\pm$ 7.69  & 238 / 1024 & 0.3512 & 0.2793 & 3.8154 \\
        llama3.2-3b & 9.69 $\pm$ 6.18   & 149 / 672  & 0.3126 & 0.3188 & 3.0083 \\
        qwen3-8b    & 20.18 $\pm$ 10.43 & 415 / 1152 & 0.4611 & 0.3668 & 4.1038 \\
        llama2-13b  & 6.20 $\pm$ 5.49   & 172 / 1600 & 0.2077 & 0.4979 & 4.8973 \\
        phi4-mini   & 6.13 $\pm$ 7.09   & 176 / 768  & 0.1845 & 0.5056 & 3.5532 \\
        \hline
    \end{tabular}
    }
    \caption{\textbf{Quantitative Statistics of Retrieval Head Dynamism.} \textbf{Mean $\pm$ Std}: Average number of dynamic retrieval heads per step and its standard deviation. \textbf{Unique / Total}: Total number of unique heads activated at least once during generation versus the total number of attention heads. \textbf{Jaccard w/ Static}: Similarity between dynamic retrieval heads and the top-20 static retrieval heads; lower values indicate fewer static heads are in the set of dynamic heads. \textbf{Adj. Jaccard}: Similarity of dynamic retrieval heads between consecutive steps. \textbf{Entropy}: Measure of distribution spread; higher values indicate broader head involvement in dynamic retrieval.}
    \label{tab:dynamism_stats}\vspace{-0.15in}
\end{table}

\begin{figure*}[t]
    \centering
    \includegraphics[width=0.9\textwidth]{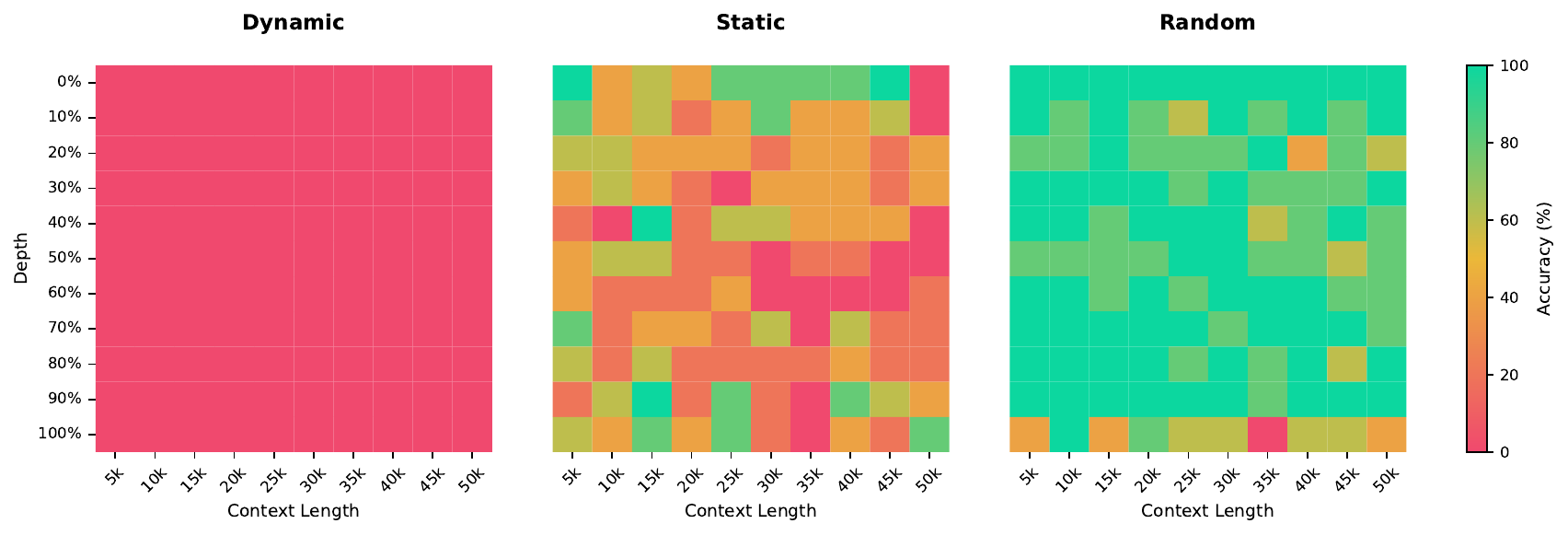}\vspace{-0.05in}
    \caption{\textbf{Impact of Head Ablation on Retrieval Performance.} Comparison of NIAH test scores after masking three different sets of attention heads: dynamic retrieval heads, top-ranked static retrieval heads, and randomly selected heads on llama3.1-8b. The x-axis shows different haystack lengths. The y-axis shows the different locations (``depth'') where the needle is inserted. The evaluation metric is Accuracy (exact string match). The average number of masked heads is kept consistent across all conditions. Masking dynamic heads (identified at each timestep via Eq.~\eqref{eq:retrieval_score_niah}) results in the most significant performance degradation, indicating their critical role in retrieval.}
    \label{fig:dynamic_masking}
\vspace{-0.08in}
\end{figure*}

There are several observations from the table: \textbf{First,} the standard deviation in the number of active heads (\textbf{Mean $\pm$ Std}) indicates that the quantity of active retrieval heads varies across timesteps. \textbf{Second,} the \textbf{Unique / Total} active heads metric demonstrates a ``long-tail'' distribution. For instance, llama3.1-8b activates 238 distinct heads for retrieval over time, which exceeds the scope of the conventional static subset (e.g., top-20). \textbf{Third,} the Jaccard similarity (\textbf{``Jaccard w/ Static''}, ranging from 0.1845 to 0.4611) indicates that only a fraction of static heads are identified as retrieval heads in a given generation step. \textbf{Fourth,} the \textbf{Adjacent Jaccard} similarity scores (ranging from 0.2793 to 0.5056) reveal a turnover rate that suggests the model frequently changes its active retrieval heads between consecutive tokens. \textbf{Finally,} the \textbf{entropy} corroborates this broad involvement. As a baseline, a uniform distribution over 20 heads yields an entropy of $\ln 20 \approx 2.99$. The observed entropy values exceed 3.0 (reaching 4.89). Together with the unique head counts, this indicates that dynamic retrieval is distributed across a wider set of heads rather than being confined to a small, fixed static subset. (See Appendix~\ref{ap:entropy_baseline} for detailed formulations).

Furthermore, the observed low similarity is robust to the choice of the static truncation threshold $k$. We conducted a sensitivity analysis by extending the static baseline to $k=50$ and $k=100$. As shown in Appendix~\ref{ap:jaccard_sensitivity}, the Jaccard similarity further decreases as $k$ increases (e.g., dropping to 0.1236 at $k=100$ for llama3.1-8b). This trend confirms that dynamic retrieval heads are not merely a rotation within a slightly larger fixed pool of heads, but rather represent a distinct and sparse distribution that remains largely uncovered even by broader static selections.


\vspace{-0.05in}
\subsection{Dynamic Retrieval Heads are Irreplaceable}\label{sec:analysis_irreplaceable}
\vspace{-0.05in}
We further claim that the dynamic retrieval heads at a specific timestep \textbf{can not be replaced by the static retrieval heads}.

\subsubsection{What will Happen without Dynamic Retrieval Heads?}

To verify Claim 2, we conducted a head ablation study. The experiment procedure is as follows:

\begin{enumerate}[leftmargin=*]\vspace{-0.1in}
    \item For each token generation step, we first execute a standard forward pass without any interventions. Discard the generated token. \vspace{-0.05in}
    \item We locate the retrieval heads at this step, as defined by Equation~\eqref{eq:retrieval_score_niah}, and label them as the set of dynamic retrieval heads. \vspace{-0.05in}
    \item We mask all the dynamic retrieval heads of this timestep, and execute a second forward pass to re-generate the token for this timestep.\vspace{-0.05in}
\end{enumerate}

For comparison, we considered two baseline approaches. We masked heads drawn from either (a) the top-ranked static retrieval heads or (b) a randomly selected set of heads. 
For fair comparison, we mask the same number of static retrieval heads/random heads as the average number of masked dynamic retrieval heads. (Details in Appendix~\ref{ap:niah_masking_setting})

This experimental design allows us to directly test our hypothesis: if dynamic retrieval heads indeed carry the primary functionality of retrieval at a given timestep, then masking them should cause a significantly greater performance degradation than masking any other set of heads. 

Figure~\ref{fig:dynamic_masking} presents the NIAH test results on \textit{meta-llama/Llama-3.1-8B-Instruct} (llama3.1-8b). The results clearly show that masking the dynamic retrieval heads leads to the most severe degradation in retrieval performance, as the colors are almost red. This far exceeds the impact of masking an equal number of static or random heads, in which only some of the pieces in the heatmap are red. 
For ROUGE-L results and other models, see Appendix~\ref{ap:niah_masking_result}.
This verifies our hypothesis that the dynamic retrieval heads are not replaceable.

\subsubsection{To What Extent do Static Retrieval Heads Help?}\label{sec:analysis_irreplaceable_k}

To further investigate the irreplaceability of dynamic retrieval heads, we also designed a progressive ablation study to analyze the model's compensatory mechanisms.

Our primary objective is to quantify the extent to which the model compensates for the loss of these optimal heads by activating additional strong static retrieval heads. To measure this, we first identify the set of \textit{compensated heads} at each timestep, i.e., heads that become retrieval heads only after $k$ dynamic heads are ablated. We then count how many of these compensated heads are in the top-20 static retrieval heads. Detailed experiment descriptions can be found in Appendix~\ref{ap:analysis_irreplaceable_k_setting}.

\begin{figure}[h]
    \centering
    \includegraphics[width=0.85\linewidth]{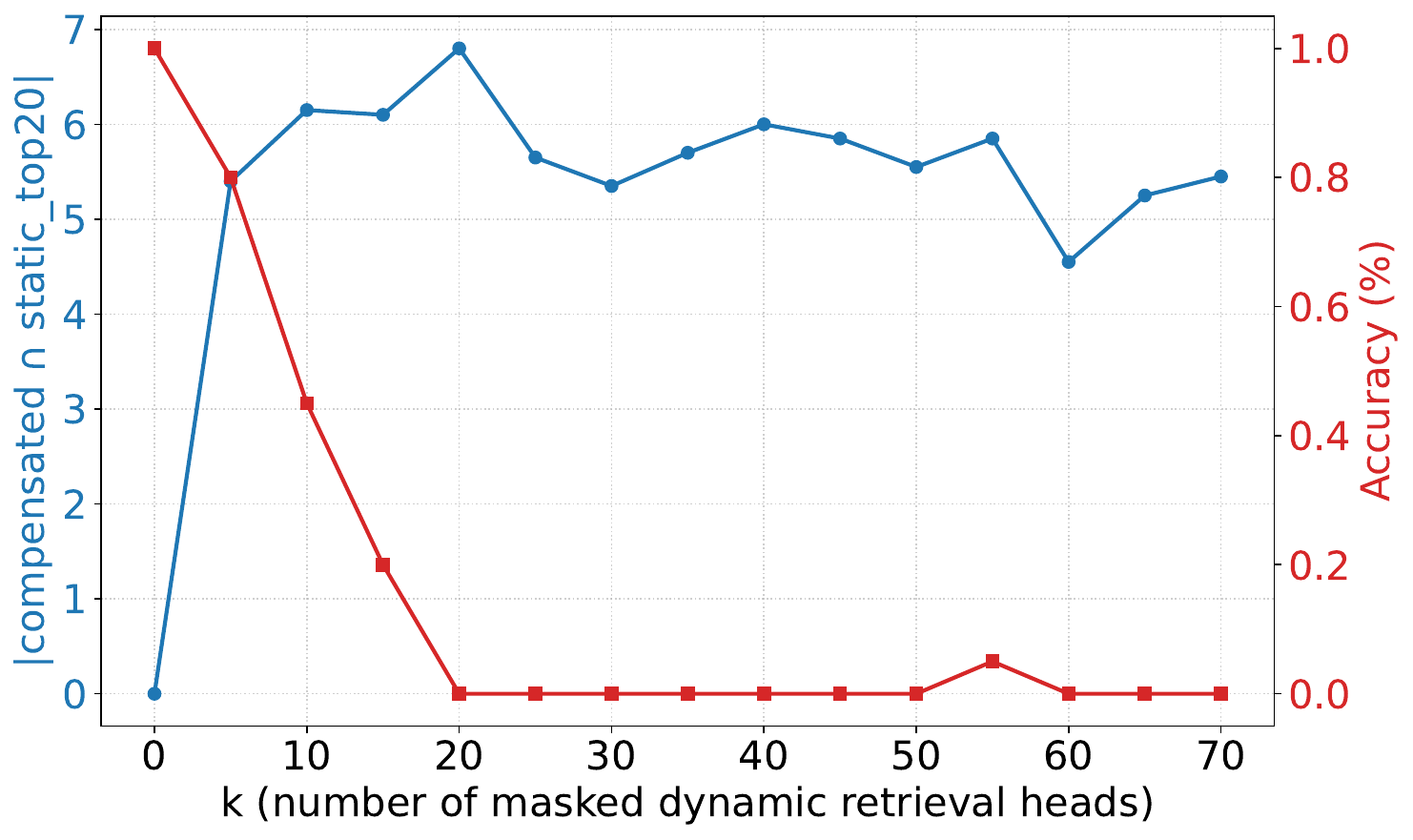}
    \caption{\textbf{Irreplaceability of Dynamic Retrieval Heads.} The plots show the degradation in NIAH performance as an increasing number ($k$) of dynamic retrieval heads are masked on llama3.1-8b. Even though the model compensates by activating top-20 static retrieval heads ({\color{blue} blue} line, left y-axis), the overall retrieval performance, measured by Accuracy ({\color{red} red} line, right y-axis), continues to decline sharply. This demonstrates that static retrieval heads cannot effectively substitute for context-specific dynamic heads.}
    \label{fig:irreplaceability}\vspace{-0.1in}
\end{figure}

Figure~\ref{fig:irreplaceability} illustrates the result on llama3.1-8b, plotting the retrieval performance ({\color{red} red} line, right y-axis) against the number of compensated heads that overlap with the top-20 static retrieval heads ({\color{blue} blue} line, left y-axis). The result reveals two critical observations. First, as the number of masked dynamic heads ($k$) increases, the model attempts to compensate by activating new heads as retrieval heads. A significant proportion of these compensated heads are indeed static retrieval heads, specifically, the overlap with the top-20 static set rises sharply to range between 4.5 and 7 for $k \geq 5$. Second, however, this compensation is insufficient. As shown by the red line, the overall retrieval performance degrades significantly. For instance, the Accuracy drops sharply from 1.0 to 0.0 as $k$ reaches 20,
suggesting that the function of dynamic retrieval heads is irreplaceable, and cannot be fully compensated for by static retrieval heads. For ROUGE-L metric and other models, see Appendix~\ref{ap:niah_irreplaceability_result}.

\subsection{Retrieval Scores are Correlated with Hidden States}\label{sec:analysis_predictable}

Our final claim is that a \textbf{strong correlation} exists between the model's hidden state and its future retrieval activations, indicating that the model employs a \textbf{predictive mechanism} for its functional behavior, specifically retrieval in this case.

To validate this, we employed Canonical Correlation Analysis (CCA) with a \textbf{temporal offset}, denoted as $k$. Specifically, we measured the linear correlation between the \textbf{final hidden state} (the output embedding of the last input token at the last layer) at timestep $n$ and the \textbf{retrieval scores} of all attention heads at a future timestep $n+k$. Detailed experiment settings can be found in Appendix~\ref{ap:cca_setting}.

\begin{figure}[h]
    \centering
    \includegraphics[width=0.85\linewidth]{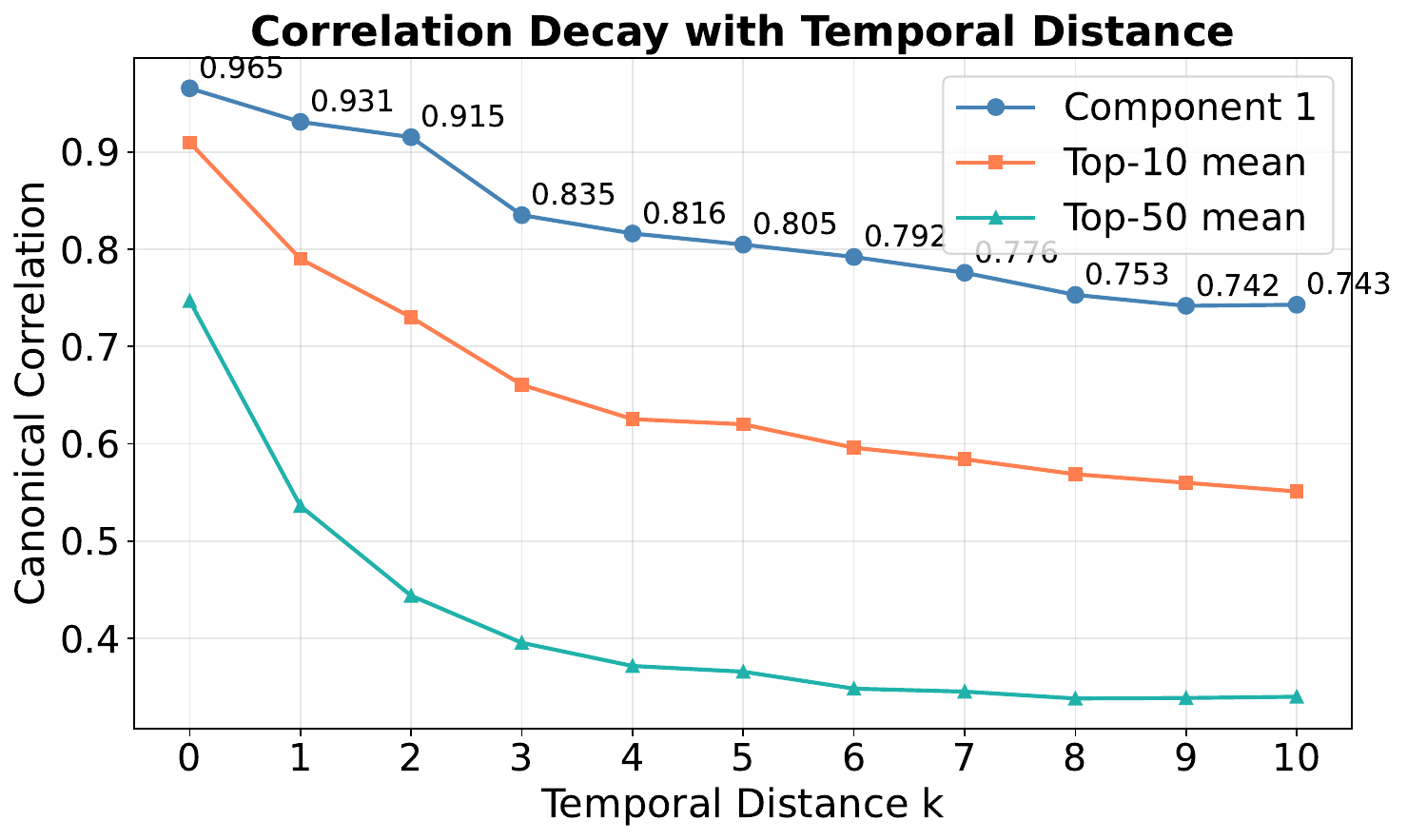}
    \caption{\textbf{Predictive Correlation Between Hidden States and Future Retrieval Scores.} Canonical Correlation Analysis (CCA) coefficients between the final hidden state at timestep $n$ and the retrieval scores at a future timestep $n+k$. The plot shows the decay of the leading (Top-1) canonical correlation, as well as the average of the Top-10 and Top-50 correlations, as the temporal offset $k$ increases. The high correlation at $k>0$ demonstrates the predictive encoding of future retrieval activations.}
    \label{fig:predictability}
\end{figure}

As shown in Figure~\ref{fig:predictability}, the canonical correlation decreases as the temporal offset $k$ increases. At $k=0$, the first canonical correlation is an exceptionally high 0.966, confirming a strong synchronous relationship between the hidden states and the retrieval scores. Besides, this correlation remains extremely high for future steps, at 0.931 for $k=1$ and 0.915 for $k=2$, indicating that the model's state acts as a predictive mechanism for retrieval operations several steps before they are executed.

While CCA demonstrates a strong linear relationship in this model, we additionally trained an MLP probe to further capture the non-linear relationships. We focused this analysis on the $k=0$ case, i.e., the retrieval scores at the same timestep as the hidden states. The probe's task was to learn the mapping from the hidden state to the specific retrieval score pattern of all heads. Detailed experiment settings can be found in Appendix~\ref{ap:mlp_probe_setting}. 

As shown in Table~\ref{tab:predictability}, the probes achieve strong performance across all models, with F1-scores ranging from 0.80 to 0.86 and AU-PRC values all exceeding 0.88, confirming that the dynamic retrieval head patterns are predictable.

\begin{table}[h]
    \centering
    \resizebox{\linewidth}{!}{
    \begin{tabular}{lcccc}
        \hline
        \textbf{LLM} & \textbf{F1} & \textbf{Precision} & \textbf{Recall} & \textbf{AUPRC} \\
        \hline
        llama3.1-8b & 0.8349 & 0.8344 & 0.8353 & 0.9173 \\
        llama3.2-3b & 0.8456 & 0.8564 & 0.8351 & 0.9289 \\
        qwen3-8b    & 0.8566 & 0.8780 & 0.8362 & 0.9339 \\
        llama2-13b  & 0.8336 & 0.8455 & 0.8220 & 0.9183 \\
        phi4-mini   & 0.8038 & 0.8219 & 0.7865 & 0.8862 \\
        \hline
    \end{tabular}
    }
    \caption{\textbf{Performance of MLP Probes in Decoding Retrieval Scores.} The probes were trained to predict retrieval head scores from the final hidden state for various LLMs. The reported metrics (Precision, Recall, F1-Score) are calculated at the optimal decision threshold, which was determined by maximizing the F1-score on the validation set's Precision-Recall curve.}
    \label{tab:predictability}
\end{table}

\vspace{-0.2in}
\section{Generalizing Dynamic Retrieval Heads in a Question Answering Task}\label{sec:hotpotqa}

In Section~\ref{sec:analysis}, we systematically investigated three core claims of retrieval heads within the controlled experimental setting of the NIAH task. However, NIAH represents a simplified ``copy-paste'' scenario where the retrieved token is directly generated. 
To consider reasoning tasks, we need a broader definition of ``retrieval'' that based on attention allocation rather than token copying.

\subsection{A Reasoning-Oriented Definition of Retrieval Score}

In complex reasoning tasks such as Question Answering, the model's behavior goes beyond simple ``copy-paste'' operations. The model must integrate multiple supporting facts to derive an answer.
Therefore, we propose a more generalized retrieval score, and correspondingly name the retrieval heads as \textbf{reasoning retrieval heads} whose retrieval score exceeds a pre-defined threshold. Based on the works of \citet{fu2024not, zhang2025query} with slight adaptation, we define the \textbf{reasoning retrieval score} $S_{\mathrm{reasoning}}(x^t, h)$ for an attention head $h$ at timestep $t$ as the proportion of attention it allocates to all supporting facts (the ``needles'') relative to the total attention it distributes across the entire effective context (excluding the interference from attention sinks~\citep{xiao2023efficient} and local attention).\footnote{Following common practice, we exclude two types of attention patterns that are not directly related to long-range retrieval: (1) \textbf{Attention Sinks}~\citep{xiao2023efficient}, where certain tokens (e.g., the initial BOS token) often receive high attention regardless of content, and (2) \textbf{Local Attention}, where heads focus on a small, fixed window of recent tokens.}
Formally,
\vspace{-0.06in}
\begin{equation}\label{eq:retrieval_score_ratio}
    S_{\mathrm{reasoning}}(x^t, h) = \frac{\sum_{i \in I_\text{needle}}\mathbf a^{h, t}_i}{\sum_{j \in I \backslash \{I_\text{sink}\cup I_\text{local}\}}\mathbf a^{h, t}_j}
\end{equation}
where $I_\text{needle}$ is the set of indices for all supporting fact tokens, while $I_\text{sink}$ and $I_\text{local}$ represent the indices of the attention sink and local attention window, respectively. Intuitively, this score relaxes the strong copy-paste condition and measures a head's focus on the correct information at a given step, better aligning with how facts are used in the LLM reasoning process.

\subsection{Experiments}\label{sec:hotpotqa_exp}

We use the HotpotQA dataset~\citep{yang2018hotpotqa} as our testbed. 
HotpotQA is a question-answering dataset that requires multi-hop reasoning, where the model must find and integrate multiple discrete supporting facts from the context to formulate a correct answer. 
To examine whether the three claims in Section \ref{sec:analysis} are still valid for the reasoning retrieval heads on the HotpotQA task, we adopt the analytical framework from Section~\ref{sec:analysis}. 

\begin{figure}[!ht]
    \centering
    \includegraphics[width=0.9\linewidth]{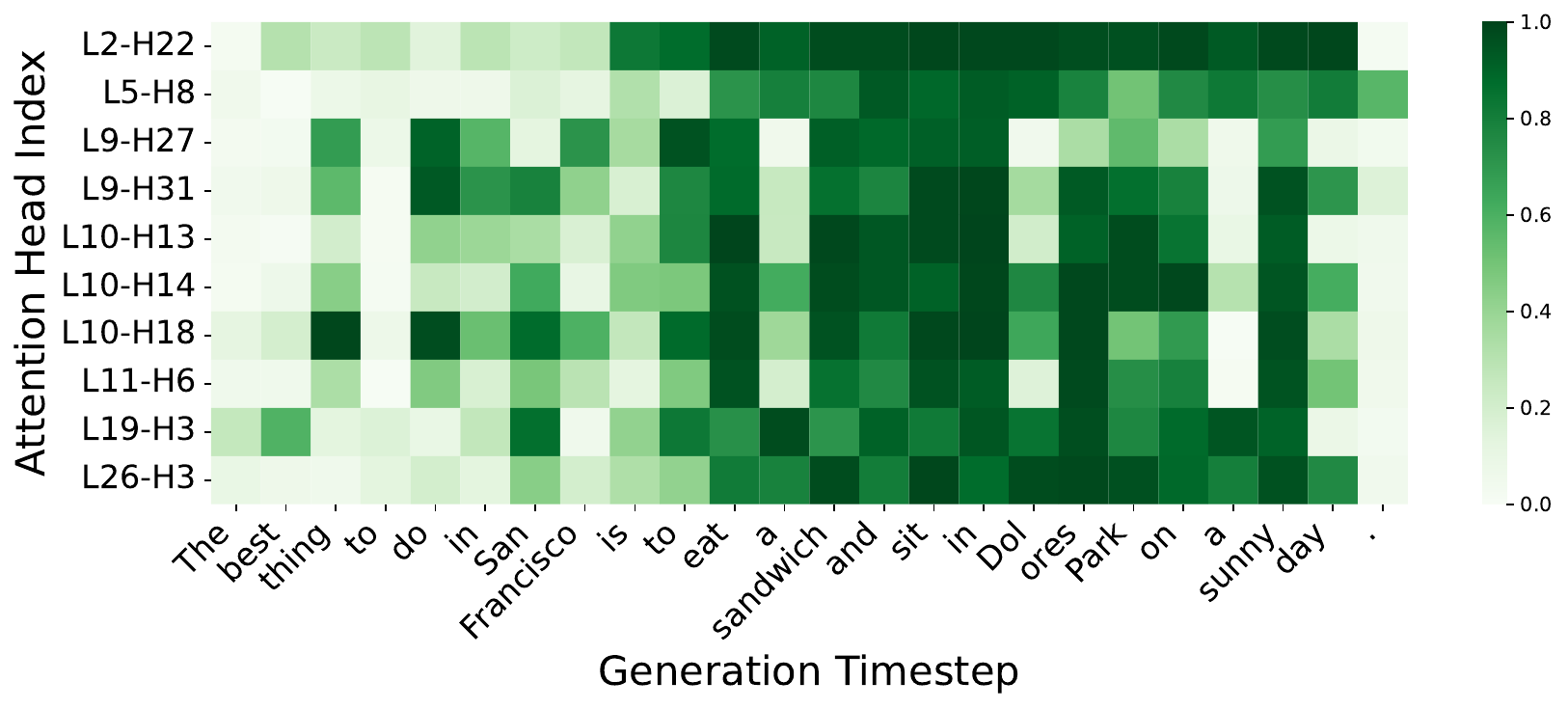}
    \caption{\textbf{Dynamic Pattern of Retrieval Heads in a Multi-Hop Reasoning Task.} The heatmap illustrates the retrieval scores (defined in Eq.~\ref{eq:retrieval_score_ratio}) for ten active retrieval heads over the course of the generation process. 
    }\vspace{-0.05in}
    \label{fig:retrieval_dynamics_hotpotqa}
\end{figure}

\paragraph{Dynamism} We first validate Claim 1. Figure~\ref{fig:retrieval_dynamics_hotpotqa} visualizes the retrieval scores of the top-10 variated retrieval heads during a generation. Consistent with the NIAH task, the retrieval head pattern is highly dynamic: no single head dominates throughout the entire process. Instead, different heads operate as retrieval heads at distinct generation stages, confirming that the dynamic nature of retrieval heads is a general phenomenon that persists in complex reasoning tasks.

\begin{figure}[!ht]
    \centering
    \includegraphics[width=0.85\linewidth]{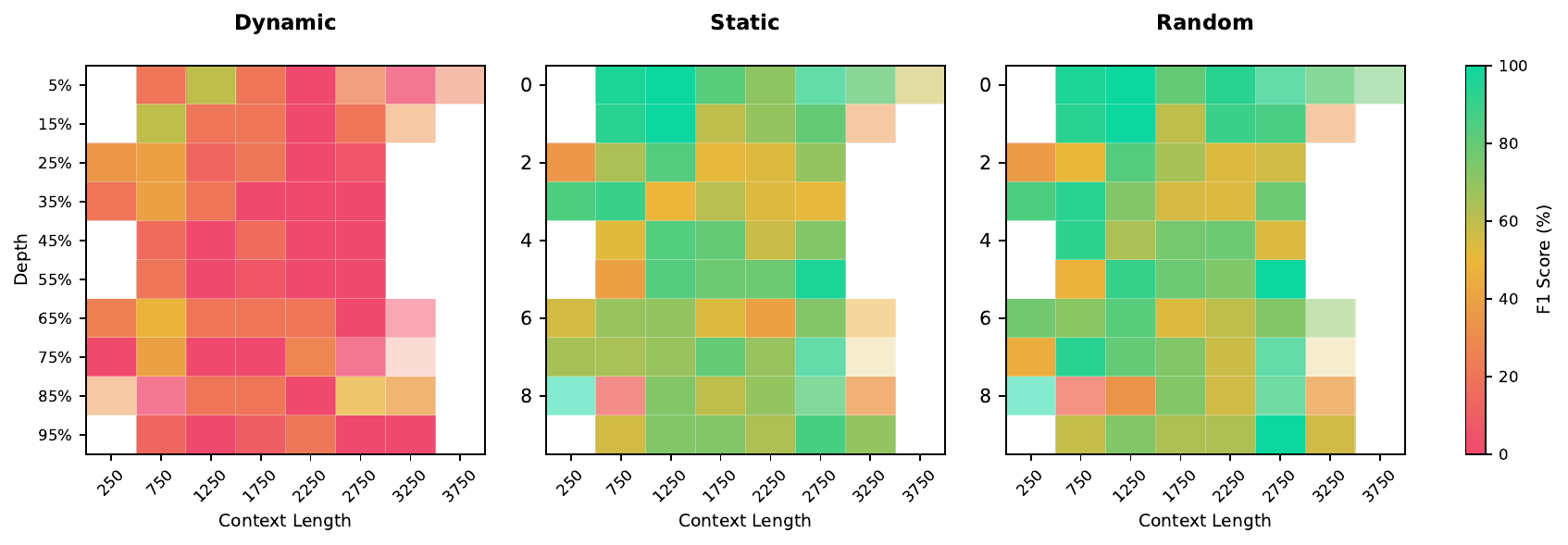}
    \caption{\textbf{Impact of Head Ablation on Multi-Hop Reasoning Performance.} F1-score comparison of HotpotQA test scores after ablating three different sets of attention heads on llama3.1-8b.
    The opacity of each cell corresponds to the number of valid samples it contains, with blank as no valid samples.
    }
    \label{fig:dynamic_masking_hotpotqa}\vspace{-0.05in}
\end{figure}

\begin{figure}[!ht]
    \centering
    \includegraphics[width=0.85\linewidth]{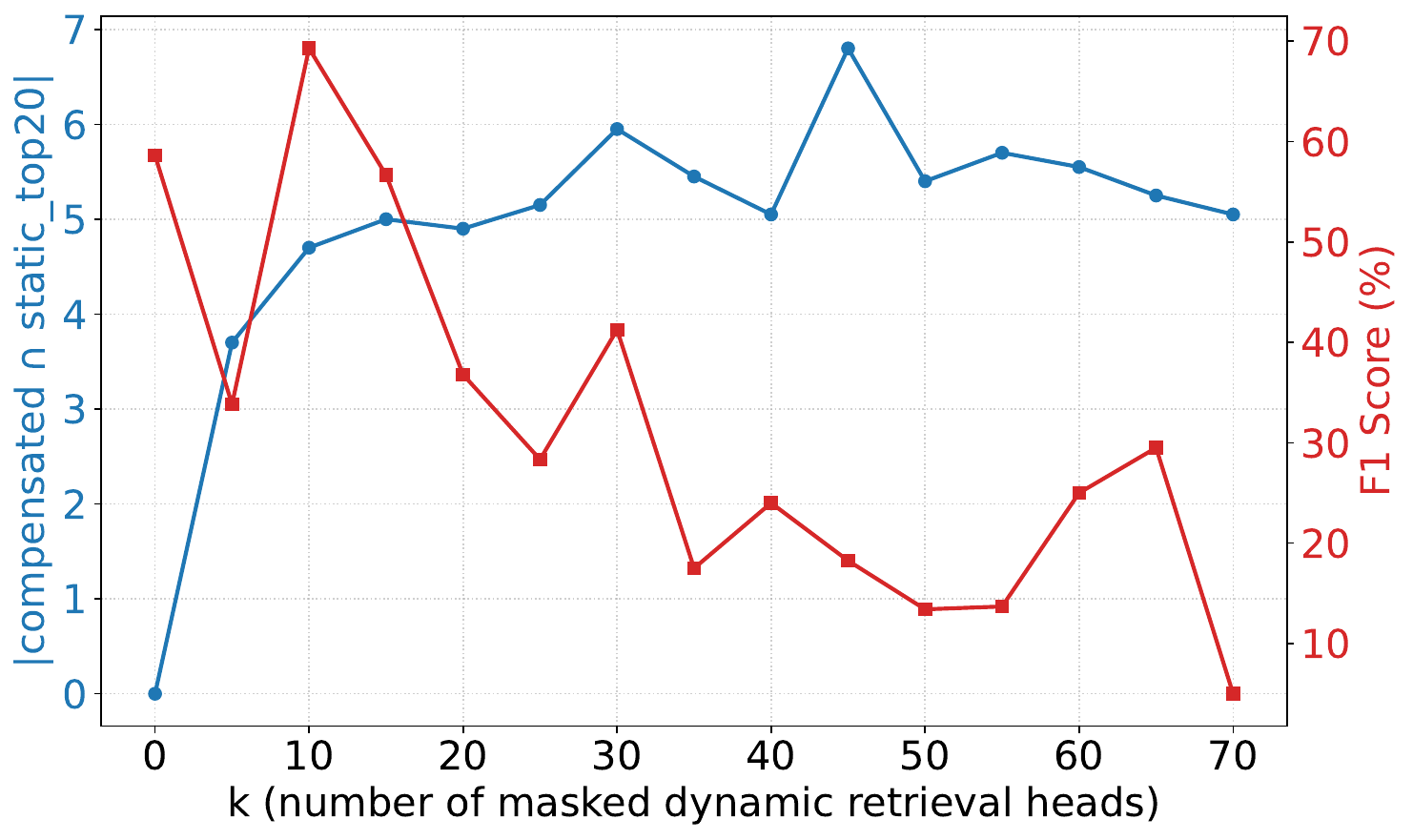}
    \caption{\textbf{Irreplaceability of Dynamic Heads in a Multi-Hop Reasoning Context.} The plots show the degradation in HotpotQA performance as an increasing number ($k$) of dynamic retrieval heads are ablated on llama3.1-8b.}\vspace{-0.15in}
    \label{fig:irreplaceability_hotpotqa}
\end{figure}

\paragraph{Irreplaceability} Next, we validate Claim 2 (Irreplaceability) through the head ablation experiment described in Section~\ref{sec:analysis_irreplaceable}. The results in both Figure~\ref{fig:dynamic_masking_hotpotqa} and Figure \ref{fig:irreplaceability_hotpotqa} show that masking the dynamic retrieval heads at each step (identified using Equation~\eqref{eq:retrieval_score_ratio}) leads to a far more severe performance degradation on the HotpotQA task than masking an equivalent number of top static retrieval heads or random heads.

\begin{figure}[h]
    \centering
    \includegraphics[width=0.85\linewidth]{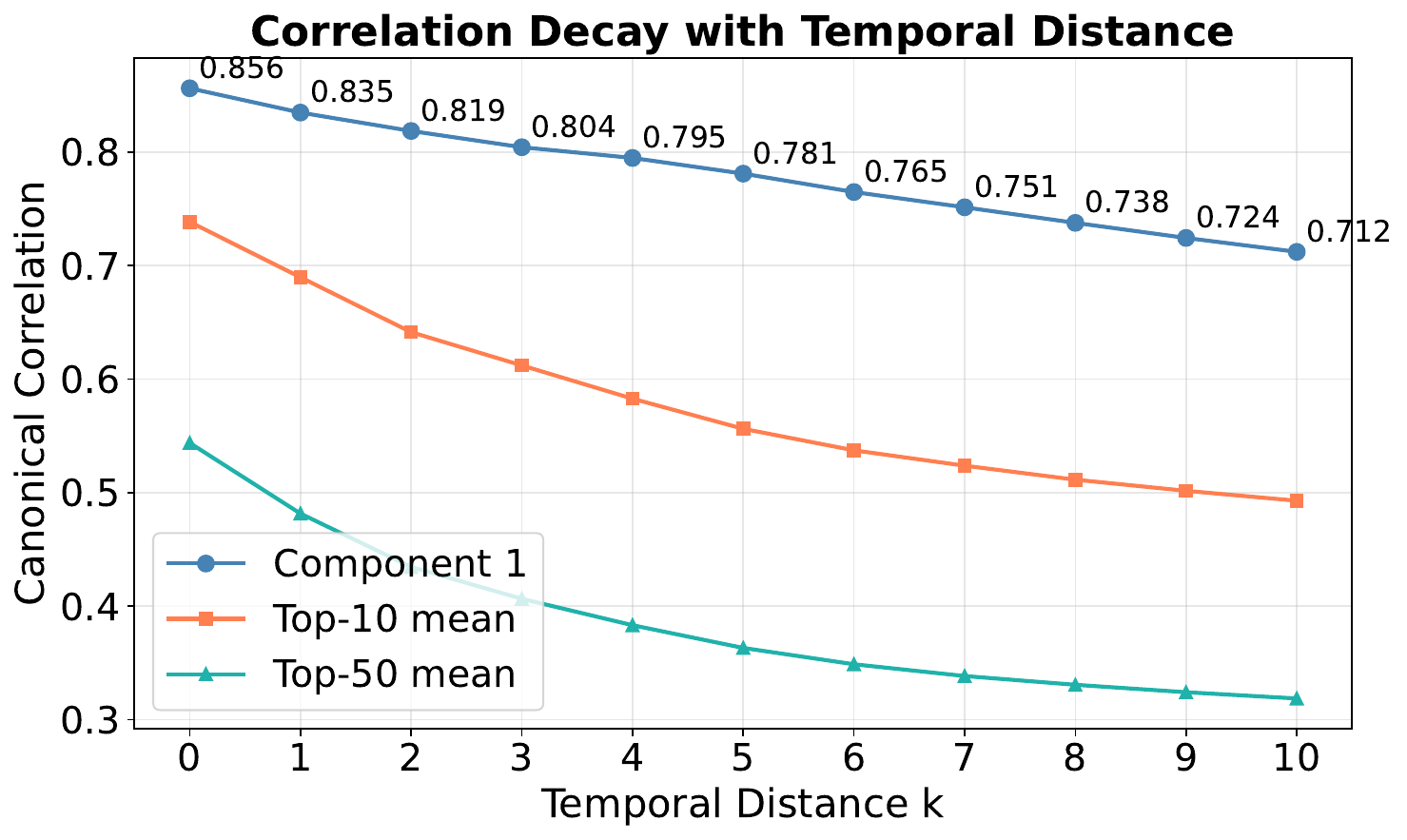}\vspace{-0.05in}
    \caption{\textbf{Temporal Decay of Correlation on a Multi-hop Reasoning Task.}}
    \label{fig:predictability_hotpotqa}\vspace{-0.1in}
\end{figure}

\paragraph{Correlation} Finally, we validate Claim 3. Using the same Temporal Offset CCA, Figure~\ref{fig:predictability_hotpotqa} shows that the strong linear correlation between hidden states and retrieval scores persists in HotpotQA. In terms of the MLP experiment, unlike the NIAH task where retrieval is binary, in HotpotQA, the retrieval score (defined in Eq.~\ref{eq:retrieval_score_ratio}) is a continuous value representing the intensity of attention on supporting facts. Consequently, we trained an MLP regressor rather than a classifier to predict the precise retrieval score vector for all heads from the final hidden state at the synchronous step ($k=0$). As shown in Table~\ref{tab:predictability_hotpotqa}, the probes achieve high $R^2$ scores (up to 0.81) across all models. This indicates that the information of retrieval scores is effectively encoded in the hidden state, verifying our claim of the strong correlation.

\begin{table}[h]
    \centering
    \resizebox{0.8\linewidth}{!}{
    \begin{tabular}{lcccc}
        \hline
        \textbf{Model} & \textbf{MSE} ($\downarrow$) & \textbf{MAE} ($\downarrow$) & \textbf{$R^2$} ($\uparrow$) \\
        \hline
        llama3.1-8b & 0.0023 & 0.0177 & 0.8120 \\
        llama3.2-3b & 0.0036 & 0.0247 & 0.8015 \\
        qwen3-8b    & 0.0050 & 0.0255 & 0.7200 \\
        llama2-13b  & 0.0009 & 0.0121 & 0.7669 \\
        phi4-mini & 0.0014 & 0.0109 & 0.7333 \\
        \hline
    \end{tabular}
    }
    \caption{\textbf{Performance of MLP Probes in Predicting Reasoning Retrieval Scores on HotpotQA.} }\vspace{-0.1in}
    \label{tab:predictability_hotpotqa}
\end{table}




\vspace{-0.05in}
\section{Case Study: Applying Dynamic Retrieval Heads to Dynamic RAG}\label{sec:experiment}

\begin{table*}[ht!]
    \centering
    \resizebox{0.9\textwidth}{!}{
    \begin{tabular}{lccccc}
        \hline
        \textbf{Model} & \textbf{Dynamic} & \textbf{Static} & \textbf{Dynamic Random} & \textbf{Fixed Random} & \textbf{w/o RAG} \\
        & EM / F1 & EM / F1 & EM / F1 & EM / F1 & EM / F1 \\
        \hline
        llama3.1-8b & \textbf{0.456 / 0.5586} & 0.398 / 0.5098 & 0.272 / 0.3670 & 0.272 / 0.3763 & 0.252 / 0.3257 \\
        llama3.2-3b & 0.384 / 0.4993 & \textbf{0.428 / 0.5386} & 0.224 / 0.3143 & 0.226 / 0.3051 & 0.184 / 0.2439 \\
        qwen3-8b & \textbf{0.286 / 0.3580} & 0.278 / 0.3429 & 0.210 / 0.2804 & 0.210 / 0.2804 & 0.220 / 0.2961 \\
        llama2-13b & \textbf{0.284 / 0.3838} & 0.278 / 0.3789 & 0.276 / 0.3762 & 0.272 / 0.3751 & 0.192 / 0.2750 \\
        phi4-mini & \textbf{0.202 / 0.2690} & 0.186 / 0.2505 & 0.082 / 0.1090 & 0.086 / 0.1111 & 0.172 / 0.2331 \\
        \hline
    \end{tabular}
    }
    \caption{Performance comparison of different retrieval strategies on the HotpotQA dataset (Exact Match (EM) / F1-Score (F1)). For each model, the better-performing strategy between Dynamic and Static is highlighted in \textbf{bold}.}
    \label{tab:dynrag}
    \vspace{-0.1in}
\end{table*}

In the previous sections, we verified that the LLM’s retrieval behavior is better characterized by dynamic retrieval heads rather than a fixed set of static retrieval heads. However, while Sections \ref{sec:analysis} and \ref{sec:hotpotqa} focus on comparing the properties of dynamic and static retrieval heads themselves, an important remaining question is how much practical advantage dynamic retrieval heads provide in real-world applications. Therefore, in this section, we designed an case study integrating them into an existing Dynamic Retrieval-Augmented Generation (Dynamic RAG) framework for question answering.

\subsection{Task Description}

Unlike traditional RAG that retrieves from external knowledge bases, retrieval heads are specialized for sourcing information already present within the model's input context. We therefore focus on an \textbf{in-context retrieval} task, evaluating the model's ability to accurately attend to and utilize key information provided in its input.

\subsection{Method}

We adapt DRAGIN~\citep{su2024dragin}, a prominent Dynamic RAG framework, for our in-context retrieval task. 
We make the following changes to integrate the retrieval heads into this framework.
At each generation step, the model's access to the context is controlled via attention masks. When no retrieval is needed, the entire context is masked, while the question and the currently generated text remain visible. When retrieval is needed, we identify the active retrieval heads for that step, determine their top-k most attended-to positions by averaging their attention scores, cluster those positions and expand a fixed-size window around each cluster. 
In the subsequent regeneration step, only the tokens corresponding to these clusters are made visible to the model via the attention mask. \footnote{We choose to use attention masking over directly rewriting the context to minimize potential disruptions to the autoregressive process, thereby isolating the impact of the retrieved information.} 
We use DRAGIN's RIND algorithm to determine when to retrieve, and the detailed pipeline can be found in Algorithm~\ref{alg:main_loop} in Appendix~\ref{ap:drag_algorithms}.

\subsection{Experiment Setup}


\paragraph{Datasets} Following Section \ref{sec:hotpotqa}, we employ the HotpotQA dataset~\citep{yang2018hotpotqa} for better practical utility evaluation compared to NIAH. 

\paragraph{Models} We selected five popular open-source models of varying sizes and architectures to ensure the generalizability of our findings: \textit{meta-llama/Llama-3.1-8B-Instruct} (llama3.1-8b), \textit{meta-llama/Llama-3.2-3B-Instruct} (llama3.2-3b), \textit{Qwen/Qwen3-8B} (qwen3-8b), \textit{meta-llama/Llama-2-13b-chat-hf} (llama2-13b), and \textit{microsoft/Phi-4-mini-instruct} (phi4-mini)~\citep{dubey2024llama,yang2025qwen3,touvron2023llama,abouelenin2025phi}. 

\paragraph{Baselines} We compare five configurations of our adapted DRAGIN framework to evaluate the efficacy of different head selection strategies:

\begin{itemize}[leftmargin=*]\vspace{-0.1in}
    \item \textbf{Dynamic}: Use dynamic retrieval heads identified by the MLP probe at each step. MLP probe from Section~\ref{sec:hotpotqa_exp} is used to predict the top-5 dynamic retrieval heads. \vspace{-0.05in}
    \item \textbf{Static}: Use 5 pre-identified static retrieval heads.\vspace{-0.05in}
    \item \textbf{Dynamic Random}: Use a new set of 5 randomly selected heads at each retrieval step.\vspace{-0.05in}
    \item \textbf{Fixed Random}: Use a fixed set of 5 randomly selected heads for the entire generation.\vspace{-0.05in}
    \item \textbf{w/o RAG}: Perform no retrieval with no context provided, relying solely on the model's parametric knowledge.
\end{itemize}


\subsection{Results}

The results are summarized in Table~\ref{tab:dynrag}. The observations indicate a superiority of dynamic retrieval heads over the other baselines:
For the majority of the tested models (llama3.1-8b, qwen3-8b, llama2-13b, and phi4-mini), the Dynamic strategy using dynamic retrieval heads achieves higher or comparable performance in both EM and F1-score compared to the Static strategy. The advantage is particularly pronounced for llama3.1-8b, where the dynamic strategy's F1-score (0.5586) is nearly 10\% higher than that of the static one (0.5098). This aligns with our findings in Sections ~\ref{sec:analysis} and~\ref{sec:hotpotqa}, suggesting that ``expert'' heads selected dynamically at each timestep can more precisely locate the information required for the current reasoning step than a fixed set of ``generalist'' heads. {An exception is the llama3.2-3b, with the Static strategy (F1=0.5386) outperforming the Dynamic one (F1=0.4993). We hypothesize this may be related to the model size: compared to other models, this model has the least number of layers with the least attention heads, indicating a possibility that each head needs to perform multiple tasks.}

To verify that the superiority of the dynamic approach is robust to the choice of the truncation threshold, we additionally evaluated the downstream performance across varying top-$k$ constraints ($k \in \{5, 10, 20\}$). As detailed in Appendix~\ref{ap:robustness_k}, the dynamic strategy consistently outperforms the static baseline across all settings, further validating the functional necessity of dynamically selecting retrieval heads.



\section{Conclusion}

This paper presents the first systematic study of retrieval heads from a dynamic perspective. Through extensive analysis on NIAH and HotpotQA tasks, we establish that retrieval head activation is highly dynamic, functionally irreplaceable, and correlated with the model's internal state. Furthermore, we demonstrate the practical utility of these insights by integrating dynamic retrieval heads into a Dynamic RAG framework, achieving significant performance gains compared to static retrieval heads. 

Beyond these immediate gains, our dynamic perspective opens up concrete directions for future research. First, our findings suggest a path toward \textbf{dynamic KV cache compression}. Existing static compression methods typically prune heads based on average importance, which risks discarding ``long-tail'' retrieval heads that are rarely active but functionally irreplaceable. Future inference systems could leverage the predictive mechanism (e.g., via lightweight probes) to dynamically retain or fetch KV pairs only for the specific heads active at the current step, achieving high compression rates while preserving sparse retrieval capabilities. Second, the token-level predictive signal can enable \textbf{hallucination detection and precision RAG}. The absence of retrieval head activation during the generation of factual claims could serve as an intrinsic, interpretable indicator of potential hallucination. Conversely, systems can use this signal to trigger external retrieval strictly when the model indicates a genuine need for information, minimizing context pollution. Ultimately, we hope this work offers a more granular understanding of LLM internal mechanisms and advances the development of efficient and interpretable model steering.

\section*{Limitations}

First, our Dynamic RAG experiment utilizes attention masking to simulate retrieval for validation purposes, rather than physically selecting and concatenating context as in standard production RAG pipelines; bridging this gap for practical deployment remains a direction for future work. Second, our method in the case study in Section~\ref{sec:experiment} relies on a learned MLP probe to predict head activation. While the probe achieves high accuracy, it is not an oracle; any prediction errors imply that the identified heads may not perfectly match the true optimal dynamic retrieval heads, potentially introducing noise into the retrieval process. Finally, our analysis primarily focuses on retrieval-intensive QA tasks (NIAH and HotpotQA); whether these findings generalize to other long-context domains, such as summarization or long-context QA, warrants further investigation.

\section*{Acknowledgement}

Yuping Lin, Pengfei He, Yingqian Cui, and Jiliang Tang are supported by the National Science Foundation (NSF) under grant numbers CNS2321416, IIS2212032, IIS2212144, IIS 2504089, DUE2234015, CNS2246050, DRL2405483 and IOS2035472.

Yue Xing is supported by NSF DMS 2515194, Open Philanthropy, NVIDIA Academic Grant Program and Google Cloud Research Credit.


\bibliography{ref}

@article{wu2024retrieval,
  title={Retrieval head mechanistically explains long-context factuality},
  author={Wu, Wenhao and Wang, Yizhong and Xiao, Guangxuan and Peng, Hao and Fu, Yao},
  journal={arXiv preprint arXiv:2404.15574},
  year={2024}
}

@article{zhang2025query,
  title={Query-Focused Retrieval Heads Improve Long-Context Reasoning and Re-ranking},
  author={Zhang, Wuwei and Yin, Fangcong and Yen, Howard and Chen, Danqi and Ye, Xi},
  journal={arXiv preprint arXiv:2506.09944},
  year={2025}
}

@article{fu2024not,
  title={Not all heads matter: A head-level kv cache compression method with integrated retrieval and reasoning},
  author={Fu, Yu and Cai, Zefan and Asi, Abedelkadir and Xiong, Wayne and Dong, Yue and Xiao, Wen},
  journal={arXiv preprint arXiv:2410.19258},
  year={2024}
}

@misc{kamradt2023niah,
  author       = {Kamradt, Greg},
  title        = {{LLMTest\_NeedleInAHaystack}: A test to measure LLM performance over long contexts},
  year         = {2023},
  publisher    = {GitHub},
  journal      = {GitHub repository},
  howpublished = {\url{https://github.com/gkamradt/LLMTest_NeedleInAHaystack}}
}

@article{olah2020zoom,
  title={Zoom in: An introduction to circuits},
  author={Olah, Chris and Cammarata, Nick and Schubert, Ludwig and Goh, Gabriel and Petrov, Michael and Carter, Shan},
  journal={Distill},
  volume={5},
  number={3},
  pages={e00024--001},
  year={2020}
}

@article{xiao2023efficient,
  title={Efficient streaming language models with attention sinks},
  author={Xiao, Guangxuan and Tian, Yuandong and Chen, Beidi and Han, Song and Lewis, Mike},
  journal={arXiv preprint arXiv:2309.17453},
  year={2023}
}

@inproceedings{lin2004rouge,
  title={Rouge: A package for automatic evaluation of summaries},
  author={Lin, Chin-Yew},
  booktitle={Text summarization branches out},
  pages={74--81},
  year={2004}
}

@article{yang2018hotpotqa,
  title={HotpotQA: A dataset for diverse, explainable multi-hop question answering},
  author={Yang, Zhilin and Qi, Peng and Zhang, Saizheng and Bengio, Yoshua and Cohen, William W and Salakhutdinov, Ruslan and Manning, Christopher D},
  journal={arXiv preprint arXiv:1809.09600},
  year={2018}
}

@article{su2024dragin,
  title={DRAGIN: dynamic retrieval augmented generation based on the information needs of large language models},
  author={Su, Weihang and Tang, Yichen and Ai, Qingyao and Wu, Zhijing and Liu, Yiqun},
  journal={arXiv preprint arXiv:2403.10081},
  year={2024}
}

@inproceedings{ridnik2021asymmetric,
  title={Asymmetric loss for multi-label classification},
  author={Ridnik, Tal and Ben-Baruch, Emanuel and Zamir, Nadav and Noy, Asaf and Friedman, Itamar and Protter, Matan and Zelnik-Manor, Lihi},
  booktitle={Proceedings of the IEEE/CVF international conference on computer vision},
  pages={82--91},
  year={2021}
}

@article{olsson2022context,
  title={In-context learning and induction heads},
  author={Olsson, Catherine and Elhage, Nelson and Nanda, Neel and Joseph, Nicholas and DasSarma, Nova and Henighan, Tom and Mann, Ben and Askell, Amanda and Bai, Yuntao and Chen, Anna and others},
  journal={arXiv preprint arXiv:2209.11895},
  year={2022}
}

@article{liu2024lost,
  title={Lost in the middle: How language models use long contexts},
  author={Liu, Nelson F and Lin, Kevin and Hewitt, John and Paranjape, Ashwin and Bevilacqua, Michele and Petroni, Fabio and Liang, Percy},
  journal={Transactions of the Association for Computational Linguistics},
  volume={12},
  pages={157--173},
  year={2024}
}

@article{li2024snapkv,
  title={Snapkv: Llm knows what you are looking for before generation},
  author={Li, Yuhong and Huang, Yingbing and Yang, Bowen and Venkitesh, Bharat and Locatelli, Acyr and Ye, Hanchen and Cai, Tianle and Lewis, Patrick and Chen, Deming},
  journal={Advances in Neural Information Processing Systems},
  volume={37},
  pages={22947--22970},
  year={2024}
}

@article{zhang2023h2o,
  title={H2o: Heavy-hitter oracle for efficient generative inference of large language models},
  author={Zhang, Zhenyu and Sheng, Ying and Zhou, Tianyi and Chen, Tianlong and Zheng, Lianmin and Cai, Ruisi and Song, Zhao and Tian, Yuandong and R{\'e}, Christopher and Barrett, Clark and others},
  journal={Advances in Neural Information Processing Systems},
  volume={36},
  pages={34661--34710},
  year={2023}
}

@article{cai2024pyramidkv,
  title={Pyramidkv: Dynamic kv cache compression based on pyramidal information funneling},
  author={Cai, Zefan and Zhang, Yichi and Gao, Bofei and Liu, Yuliang and Li, Yucheng and Liu, Tianyu and Lu, Keming and Xiong, Wayne and Dong, Yue and Hu, Junjie and others},
  journal={arXiv preprint arXiv:2406.02069},
  year={2024}
}

@article{vaswani2017attention,
  title={Attention is all you need},
  author={Vaswani, Ashish and Shazeer, Noam and Parmar, Niki and Uszkoreit, Jakob and Jones, Llion and Gomez, Aidan N and Kaiser, {\L}ukasz and Polosukhin, Illia},
  journal={Advances in neural information processing systems},
  volume={30},
  year={2017}
}

@article{radford2019language,
  title={Language models are unsupervised multitask learners},
  author={Radford, Alec and Wu, Jeffrey and Child, Rewon and Luan, David and Amodei, Dario and Sutskever, Ilya and others},
  journal={OpenAI blog},
  volume={1},
  number={8},
  pages={9},
  year={2019}
}

@article{brown2020language,
  title={Language models are few-shot learners},
  author={Brown, Tom and Mann, Benjamin and Ryder, Nick and Subbiah, Melanie and Kaplan, Jared D and Dhariwal, Prafulla and Neelakantan, Arvind and Shyam, Pranav and Sastry, Girish and Askell, Amanda and others},
  journal={Advances in neural information processing systems},
  volume={33},
  pages={1877--1901},
  year={2020}
}

@article{chowdhery2023palm,
  title={Palm: Scaling language modeling with pathways},
  author={Chowdhery, Aakanksha and Narang, Sharan and Devlin, Jacob and Bosma, Maarten and Mishra, Gaurav and Roberts, Adam and Barham, Paul and Chung, Hyung Won and Sutton, Charles and Gehrmann, Sebastian and others},
  journal={Journal of Machine Learning Research},
  volume={24},
  number={240},
  pages={1--113},
  year={2023}
}

@article{hoffmann2022training,
  title={Training compute-optimal large language models},
  author={Hoffmann, Jordan and Borgeaud, Sebastian and Mensch, Arthur and Buchatskaya, Elena and Cai, Trevor and Rutherford, Eliza and Casas, Diego de Las and Hendricks, Lisa Anne and Welbl, Johannes and Clark, Aidan and others},
  journal={arXiv preprint arXiv:2203.15556},
  year={2022}
}

@article{touvron2023llama,
  title={Llama 2: Open foundation and fine-tuned chat models},
  author={Touvron, Hugo and Martin, Louis and Stone, Kevin and Albert, Peter and Almahairi, Amjad and Babaei, Yasmine and Bashlykov, Nikolay and Batra, Soumya and Bhargava, Prajjwal and Bhosale, Shruti and others},
  journal={arXiv preprint arXiv:2307.09288},
  year={2023}
}

@article{garg2022can,
  title={What can transformers learn in-context? a case study of simple function classes},
  author={Garg, Shivam and Tsipras, Dimitris and Liang, Percy S and Valiant, Gregory},
  journal={Advances in neural information processing systems},
  volume={35},
  pages={30583--30598},
  year={2022}
}

@article{xie2021explanation,
  title={An explanation of in-context learning as implicit bayesian inference},
  author={Xie, Sang Michael and Raghunathan, Aditi and Liang, Percy and Ma, Tengyu},
  journal={arXiv preprint arXiv:2111.02080},
  year={2021}
}

@inproceedings{press2023measuring,
  title={Measuring and narrowing the compositionality gap in language models},
  author={Press, Ofir and Zhang, Muru and Min, Sewon and Schmidt, Ludwig and Smith, Noah A and Lewis, Mike},
  booktitle={Findings of the Association for Computational Linguistics: EMNLP 2023},
  pages={5687--5711},
  year={2023}
}

@article{voita2019analyzing,
  title={Analyzing multi-head self-attention: Specialized heads do the heavy lifting, the rest can be pruned},
  author={Voita, Elena and Talbot, David and Moiseev, Fedor and Sennrich, Rico and Titov, Ivan},
  journal={arXiv preprint arXiv:1905.09418},
  year={2019}
}

@article{michel2019sixteen,
  title={Are sixteen heads really better than one?},
  author={Michel, Paul and Levy, Omer and Neubig, Graham},
  journal={Advances in neural information processing systems},
  volume={32},
  year={2019}
}

@article{elhage2021mathematical,
  title={A mathematical framework for transformer circuits},
  author={Elhage, Nelson and Nanda, Neel and Olsson, Catherine and Henighan, Tom and Joseph, Nicholas and Mann, Ben and Askell, Amanda and Bai, Yuntao and Chen, Anna and Conerly, Tom and others},
  journal={Transformer Circuits Thread},
  volume={1},
  number={1},
  pages={12},
  year={2021}
}

@article{elhage2022toy,
  title={Toy models of superposition},
  author={Elhage, Nelson and Hume, Tristan and Olsson, Catherine and Schiefer, Nicholas and Henighan, Tom and Kravec, Shauna and Hatfield-Dodds, Zac and Lasenby, Robert and Drain, Dawn and Chen, Carol and others},
  journal={arXiv preprint arXiv:2209.10652},
  year={2022}
}

@article{wang2022interpretability,
  title={Interpretability in the wild: a circuit for indirect object identification in gpt-2 small},
  author={Wang, Kevin and Variengien, Alexandre and Conmy, Arthur and Shlegeris, Buck and Steinhardt, Jacob},
  journal={arXiv preprint arXiv:2211.00593},
  year={2022}
}

@article{lewis2020retrieval,
  title={Retrieval-augmented generation for knowledge-intensive nlp tasks},
  author={Lewis, Patrick and Perez, Ethan and Piktus, Aleksandra and Petroni, Fabio and Karpukhin, Vladimir and Goyal, Naman and K{\"u}ttler, Heinrich and Lewis, Mike and Yih, Wen-tau and Rockt{\"a}schel, Tim and others},
  journal={Advances in neural information processing systems},
  volume={33},
  pages={9459--9474},
  year={2020}
}

@inproceedings{borgeaud2022improving,
  title={Improving language models by retrieving from trillions of tokens},
  author={Borgeaud, Sebastian and Mensch, Arthur and Hoffmann, Jordan and Cai, Trevor and Rutherford, Eliza and Millican, Katie and Van Den Driessche, George Bm and Lespiau, Jean-Baptiste and Damoc, Bogdan and Clark, Aidan and others},
  booktitle={International conference on machine learning},
  pages={2206--2240},
  year={2022},
  organization={PMLR}
}

@article{press2021train,
  title={Train short, test long: Attention with linear biases enables input length extrapolation},
  author={Press, Ofir and Smith, Noah A and Lewis, Mike},
  journal={arXiv preprint arXiv:2108.12409},
  year={2021}
}

@article{su2024roformer,
  title={Roformer: Enhanced transformer with rotary position embedding},
  author={Su, Jianlin and Ahmed, Murtadha and Lu, Yu and Pan, Shengfeng and Bo, Wen and Liu, Yunfeng},
  journal={Neurocomputing},
  volume={568},
  pages={127063},
  year={2024},
  publisher={Elsevier}
}

@article{yang2025qwen3,
  title={Qwen3 technical report},
  author={Yang, An and Li, Anfeng and Yang, Baosong and Zhang, Beichen and Hui, Binyuan and Zheng, Bo and Yu, Bowen and Gao, Chang and Huang, Chengen and Lv, Chenxu and others},
  journal={arXiv preprint arXiv:2505.09388},
  year={2025}
}

@article{abouelenin2025phi,
  title={Phi-4-mini technical report: Compact yet powerful multimodal language models via mixture-of-loras},
  author={Abouelenin, Abdelrahman and Ashfaq, Atabak and Atkinson, Adam and Awadalla, Hany and Bach, Nguyen and Bao, Jianmin and Benhaim, Alon and Cai, Martin and Chaudhary, Vishrav and Chen, Congcong and others},
  journal={arXiv preprint arXiv:2503.01743},
  year={2025}
}

@article{dubey2024llama,
  title={The llama 3 herd of models},
  author={Dubey, Abhimanyu and Jauhri, Abhinav and Pandey, Abhinav and Kadian, Abhishek and Al-Dahle, Ahmad and Letman, Aiesha and Mathur, Akhil and Schelten, Alan and Yang, Amy and Fan, Angela and others},
  journal={arXiv preprint arXiv:2407.21783},
  year={2024}
}

\appendix

\section{Experimental Setup for Needle-in-a-Haystack Masking}\label{ap:niah_masking}

For reproducibility, all experiments in this paper use greedy sampling as the LLM decoding strategy. Each experiment can be perform on a single NVIDIA H200 GPU.

\subsection{Needle-in-a-Haystack Test Setting}\label{ap:niah_masking_setting}

The original Needle-in-a-Haystack (NIAH) test~\citep{kamradt2023niah} was proposed in 2023, while most of the models we study in this work were released after this. This raises the possibility that these models were exposed to the NIAH data during their training phase. To prevent potential data leakage, we use \textbf{dynamically, randomly generated UUID strings} as the needle string as a substitute, to exclude the possibility that the model has seen the Needle data during its training phase. Its specific format is as follows:

\begin{itemize}
    \item Needle: The magic word is [UUID].
    \item Question: What is the magic word?
\end{itemize}


For the experiment conducted in Section~\ref{sec:analysis_irreplaceable}, the detailed experiment setting is as follows:

For evaluation criteria, we use two types: accuracy and ROUGE-L~\citep{lin2004rouge}. Accuracy is to check whether the model's response completely contains the UUID string. If it does, the score is 1.0; otherwise, it is 0.0, even if there is only a one-character difference.

We extend the context length to the specified length by repeatedly concatenating and truncating the haystack text. Then, we randomly select a depth, backtrack to the nearest end of a sentence to insert the needle, and apply a dialogue template to construct the input. The dialogue template is as follows: 

\begin{tcolorbox}[
    breakable,
    title=NIAH Test Prompt,
    colback=blue!5!white,
    colframe=blue!75!black,
    fonttitle=\bfseries,
]
\textbf{System}

You are a helpful AI bot that answers questions for a user. Keep your response short and direct.

\textbf{User}

Context:

\{A Haystack with a Needle inserted in\}

Question:

\{Question\}

Instruction:

Don't give information outside the document or repeat your findings.
\end{tcolorbox}

To obtain the data for Figure \ref{fig:dynamic_masking}, we conducted 5 independent runs for each grid cell and reported the average metric value.

To collect the data for Figure~\ref{fig:irreplaceability}, we conducted 20 independent runs at intervals of $k=5$, with the haystack length fixed at 5000 tokens. The reported results are the averages of these trials.

\subsection{Detailed Setting for the Experiment in Section~\ref{sec:analysis_irreplaceable_k}}\label{ap:analysis_irreplaceable_k_setting}


In the ablation study involving the masking of $k$ dynamic heads, our goal is to quantify the model's attempt to compensate using static retrieval heads. To do this, we track the compensated heads and measure their overlap with the Top-20 static retrieval heads.

A key methodological challenge is how to aggregate this metric over a full generation sequence. We observed that under heavy ablation (i.e., large $k$), the model's retrieval mechanism often collapses in later generation timesteps, resulting in zero retrieval heads. Consequently, a simple average across all timesteps would include these zero-values, artificially deflating the metric and failing to reflect the model's actual capacity to utilize compensated heads. To address this and robustly capture the model's peak compensatory effort, we record the maximum number of compensated heads observed at any single timestep within each sample's generation.

Formally, for each sample $s$, let $H_{s, t}$ be the set of dynamic retrieval heads at timestep $t$ before masking, let $H'_{s, t}$ be the set of dynamic retrieval heads at timestep $t$ after masking. Let $E_{s, t} = H'_{s, t} - H_{s, t}$ be the set of compensated heads at timestep $t$. We compute the maximum intersection with the top-20 static set, $H_{\text{static}}$, within the sample $s$: $m_s = \max_t |E_{s, t} \cap H_{\text{static}}|$. The final metric is the average of $m_s$ over all samples.

\subsection{Details for Entropy Metric in Dynamism Analysis}\label{ap:entropy_baseline}

We calculate the entropy of the retrieval score distribution to measure how broadly retrieval responsibility is shared. Let $p_h$ be the probability that head $h$ is activated as a retrieval head across all timesteps. The entropy is defined as:

\begin{equation} S = - \sum_{h} p_h \ln p_h \end{equation}

To provide a baseline for interpretation, consider a scenario where retrieval is exclusively and uniformly performed by the top-20 static heads. In this case, $p_h = \frac{1}{20}$ or these 20 heads and 0 for others. The resulting entropy would be:

\begin{equation} S_{baseline} = - \sum_{i=1}^{20} \frac{1}{20} \ln \left(\frac{1}{20}\right) = \ln(20) \approx 2.9957 \end{equation}

Our observed entropy values are consistently higher than this baseline (e.g., 3.8154 for llama3.1-8b). Combining with the low Jaccard w/ Static values, this indicates that the effective number of heads participating in retrieval is significantly larger than 20.

\subsection{Experimental Settings for Canonical Correlation Analysis}\label{ap:cca_setting}

To analyze the correlation between the model's final hidden states and retrieval scores, we employed Canonical Correlation Analysis (CCA). The detailed experimental procedure is as follows:

\paragraph{Data Preprocessing} We first standardized the retrieval scores (min 0, max 1) to ensure consistent scaling across attention heads. To improve computational efficiency and focus on the principal signal subspaces, we applied Principal Component Analysis (PCA) to both the hidden states and retrieval scores prior to CCA. We retained principal components explaining 95\% of the variance for the hidden states and 99\% for the retrieval scores.

\paragraph{CCA Configuration} We set the number of canonical components to 50.

\paragraph{Temporal Offset} We analyzed the correlation with a temporal offset $k$ ranging from 0 to 10. For each offset $k$, we paired the hidden state at timestep $n$ with the retrieval scores at timestep $n+k$. Any samples where $n+k$ exceeded the sequence length were excluded from the analysis.

\subsection{Experimental Settings for MLP Probe Training} \label{ap:mlp_probe_setting}

To investigate the fine-grained correlation of retrieval head patterns, we trained a Multi-Layer Perceptron (MLP) probe for each LLM. The detailed configuration is as follows:

\paragraph{Model Architecture} The probe is a feed-forward neural network consisting of three hidden layers with dimensions [8192,4096,4096]. We applied a dropout rate of 0.1 after each hidden layer to prevent overfitting. The input dimension matches the hidden size of the respective LLM, and the output dimension corresponds to the total number of attention heads.

\paragraph{Training Configuration} The models were trained for 100 epochs with a batch size of 128. We used the Adam optimizer with a learning rate of $3 \times 10^{-4}$ and a scheduler that reduces the learning rate upon a plateau in validation loss (patience set to 3 epochs). To handle the sparse retrieval score distribution of attention heads, we employed the Asymmetric Loss~\citep{ridnik2021asymmetric} as the objective function. Gradients were clipped at a maximum norm of 1.0 to ensure stability.

\paragraph{Data Split} The dataset collected from the NIAH runs was split into training (70\%), validation (20\%), and testing (10\%) sets. All reported metrics (Precision, Recall, F1, AUPRC) are evaluated on the held-out test set.

\section{Sensitivity Analysis of Jaccard Similarity}
\label{ap:jaccard_sensitivity}

To address potential concerns regarding the arbitrary choice of the truncation threshold ($k=20$) for identifying static retrieval heads, we report the Jaccard similarity between dynamic retrieval heads and static heads at varying thresholds ($k \in \{20, 50, 100\}$). 

Table~\ref{tab:jaccard_k_sensitivity} summarizes the results across all evaluated models. We observe a consistent decrease in Jaccard similarity as $k$ increases. This indicates that the expansion of the static set (the denominator in Jaccard similarity) significantly outpaces the inclusion of additional dynamic retrieval heads in the intersection. This result reinforces our finding that dynamic retrieval heads originate from a broad, long-tail distribution of attention heads rather than being confined to a slightly expanded static subset.

\begin{table}[h]
\centering
\resizebox{1.0\linewidth}{!}{
\begin{tabular}{lccc}
\hline
Model & Jaccard ($k=20$) & Jaccard ($k=50$) & Jaccard ($k=100$) \\
\hline
Llama-3.1-8B & 0.3512 & 0.2106 & 0.1236 \\
Llama-3.2-3B & 0.3126 & 0.1755 & 0.0964 \\
Qwen3-8B     & 0.4611 & 0.3011 & 0.1799 \\
Llama-2-13B  & 0.2077 & 0.1116 & 0.0612 \\
Phi-4-mini   & 0.1845 & 0.1087 & 0.0606 \\
\hline
\end{tabular}
}
\caption{Jaccard similarity between dynamic retrieval heads and static retrieval heads at different truncation thresholds ($k$).}
\label{tab:jaccard_k_sensitivity}
\end{table}

\section{Robustness of Dynamic Retrieval Across Different $k$ Thresholds}
\label{ap:robustness_k}

In our primary Dynamic RAG case study (Section \ref{sec:experiment}), we demonstrated the effectiveness of dynamically selecting retrieval heads. To further ensure that this advantage is not an artifact of a specific threshold, we conducted a robustness check on the llama3.1-8b model by varying the number of allowed retrieval heads, denoted as $k \in \{5, 10, 20\}$. 

Table~\ref{tab:robustness_k} presents the Exact Match (EM) and F1 scores across the dynamic, static, and random baseline strategies. We observe two key phenomena:

\textbf{First, consistent superiority.} The dynamic strategy consistently outperforms both the static baseline and the random baselines across all evaluated $k$ values. This confirms that selecting heads based on the current timestep's predictive state is fundamentally more effective than relying on a fixed, historically aggregated static set.

\textbf{Second, the sparsity constraint.} Counter-intuitively, the absolute performance for all methods drops as $k$ increases from $5$ to $10$ and $20$. This phenomenon aligns perfectly with our statistical findings in Section \ref{sec:analysis_dynamism}. As shown in Table \ref{tab:dynamism_stats}, the average number of active dynamic retrieval heads per step for llama3.1-8b is relatively small and highly sparse (e.g., concentrated in a few heads at any exact moment). Forcing the model to utilize a strict top-10 or top-20 heads introduces ``noise injection.'' In other words, when $k$ exceeds the actual number of required retrieval heads for a specific step, the system artificially forces non-retrieval heads to participate in the retrieval operation, which interferes with the generation and degrades the overall downstream accuracy. This further highlights the necessity of dynamic sparsity over static inclusion.

\begin{table}[h]
\centering
\resizebox{1.0\linewidth}{!}{
\begin{tabular}{lccc}
\hline
Method & $k=5$ (EM/F1) & $k=10$ (EM/F1) & $k=20$ (EM/F1) \\
\hline
Dynamic (Ours) & \textbf{0.456 / 0.5586} & \textbf{0.286 / 0.3877} & \textbf{0.292 / 0.3967} \\
Static & 0.398 / 0.5098 & 0.280 / 0.3767 & 0.282 / 0.3810 \\
Dynamic Random & 0.272 / 0.3670 & 0.280 / 0.3807 & 0.274 / 0.3713 \\
Fixed Random   & 0.272 / 0.3763 & 0.276 / 0.3711 & 0.276 / 0.3751 \\
\hline
\end{tabular}
}
\caption{Downstream QA performance (EM/F1) on llama3.1-8b under varying top-$k$ constraints. Bold values indicate the best performance in each column.}
\label{tab:robustness_k}
\end{table}

\section{Additional Experiment Results}

\subsection{All Heads Ablation Results}\label{ap:niah_masking_result}

\begin{figure*}[t]
    \centering
    \includegraphics[width=0.9\textwidth]{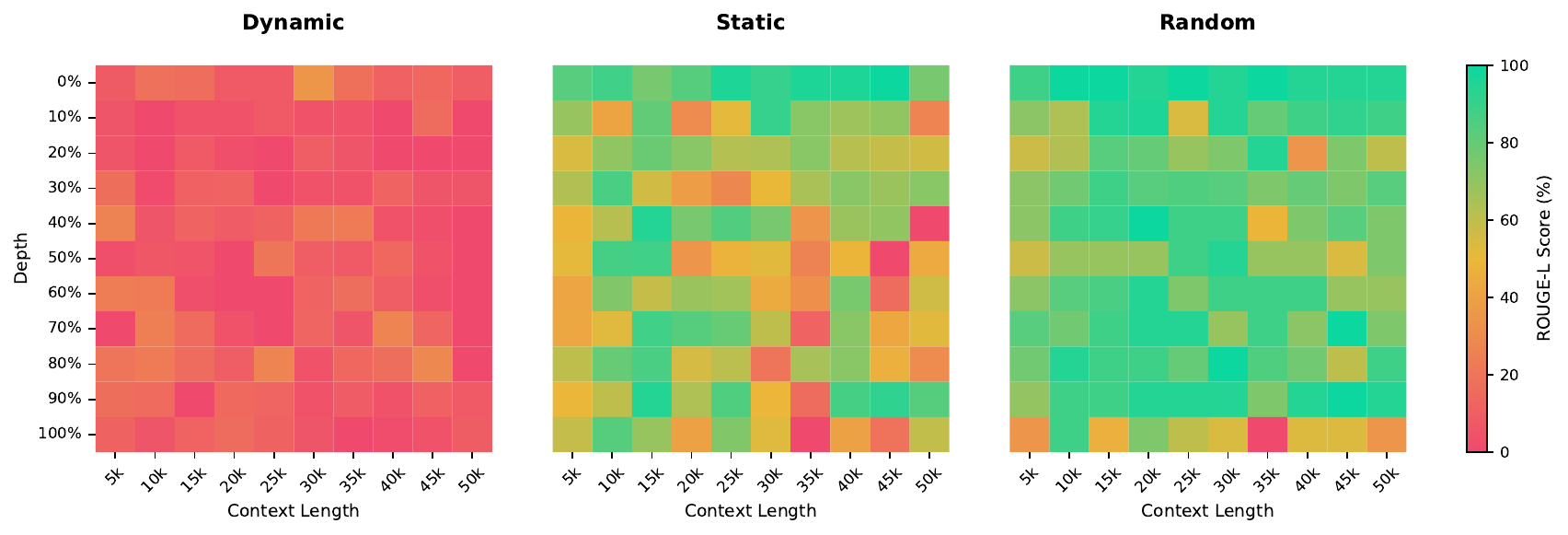}
    \caption{Head Ablation on NIAH test on llama3.1-8b. Using ROUGE-L as the metric.}
    \label{fig:dynamic_masking_rouge}
\end{figure*}

\begin{figure*}
    \centering

    \begin{subfigure}{\textwidth}
        \centering
        \includegraphics[width=0.9\textwidth]{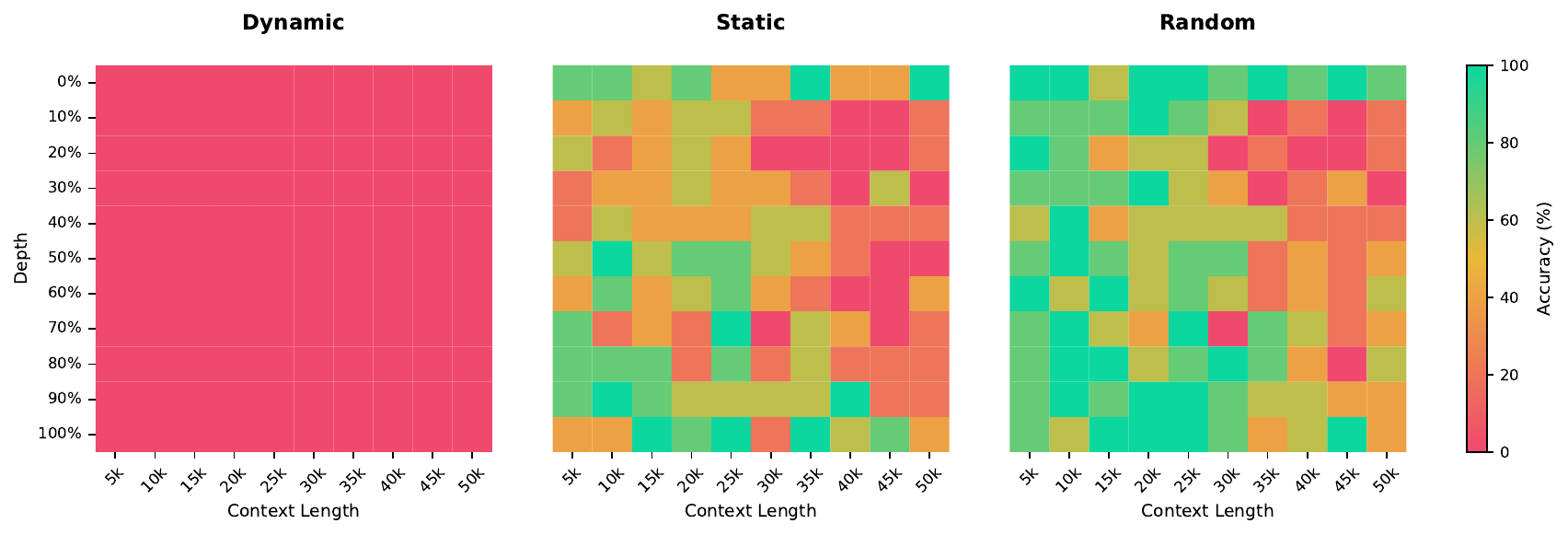}
        \caption{Accuracy}
    \end{subfigure}
    
    \begin{subfigure}{\textwidth}
        \centering
        \includegraphics[width=0.9\textwidth]{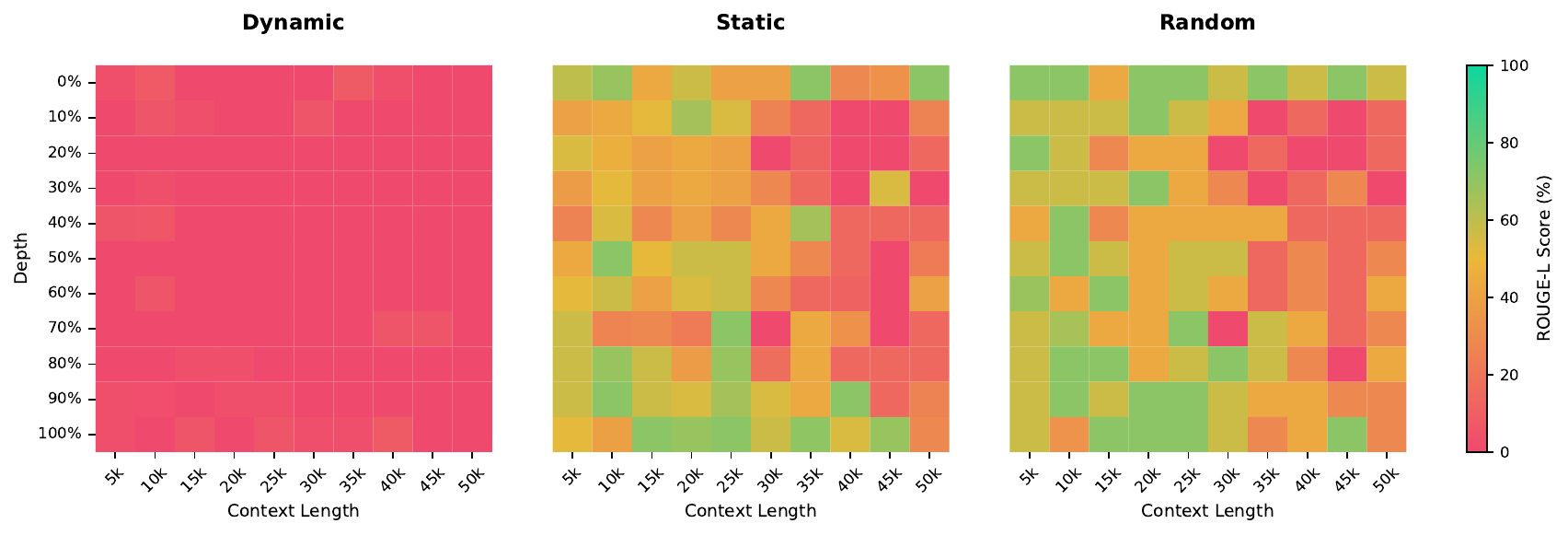}
        \caption{ROUGE-L}
    \end{subfigure}

    \caption{Head Ablation on NIAH test on llama3.2-3b.}
    \label{fig:ap_masking_llama3.2-3b}
\end{figure*}

\begin{figure*}
    \centering

    \begin{subfigure}{\textwidth}
        \centering
        \includegraphics[width=0.9\textwidth]{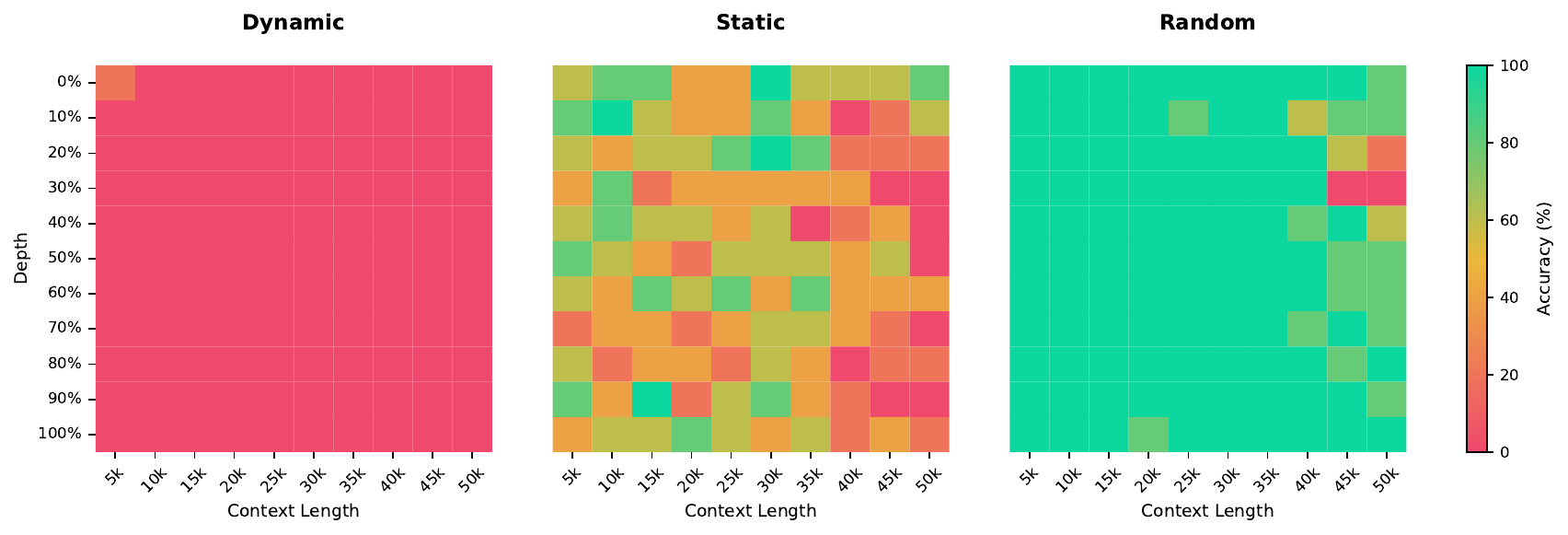}
        \caption{Accuracy}
    \end{subfigure}
    
    \begin{subfigure}{\textwidth}
        \centering
        \includegraphics[width=0.9\textwidth]{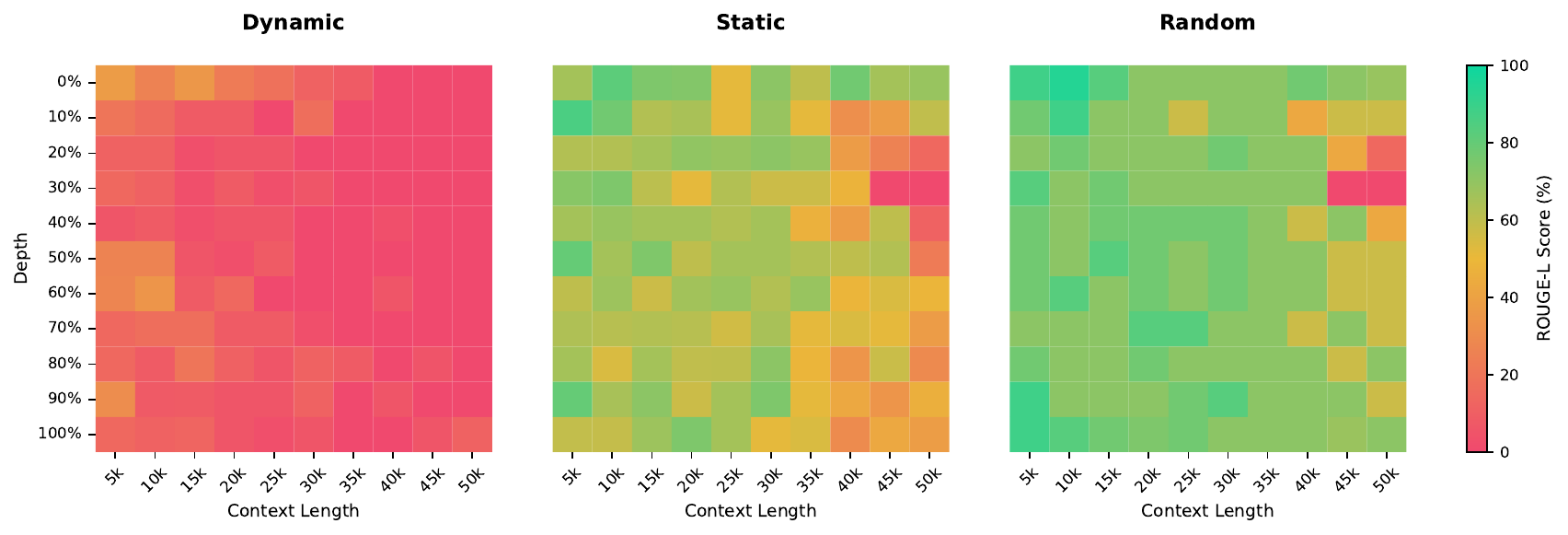}
        \caption{ROUGE-L}
    \end{subfigure}

    \caption{Head Ablation on NIAH test on qwen3-8b.}
    \label{fig:ap_masking_qwen3-8b}
\end{figure*}

\begin{figure*}
    \centering
    \includegraphics[width=0.9\textwidth]{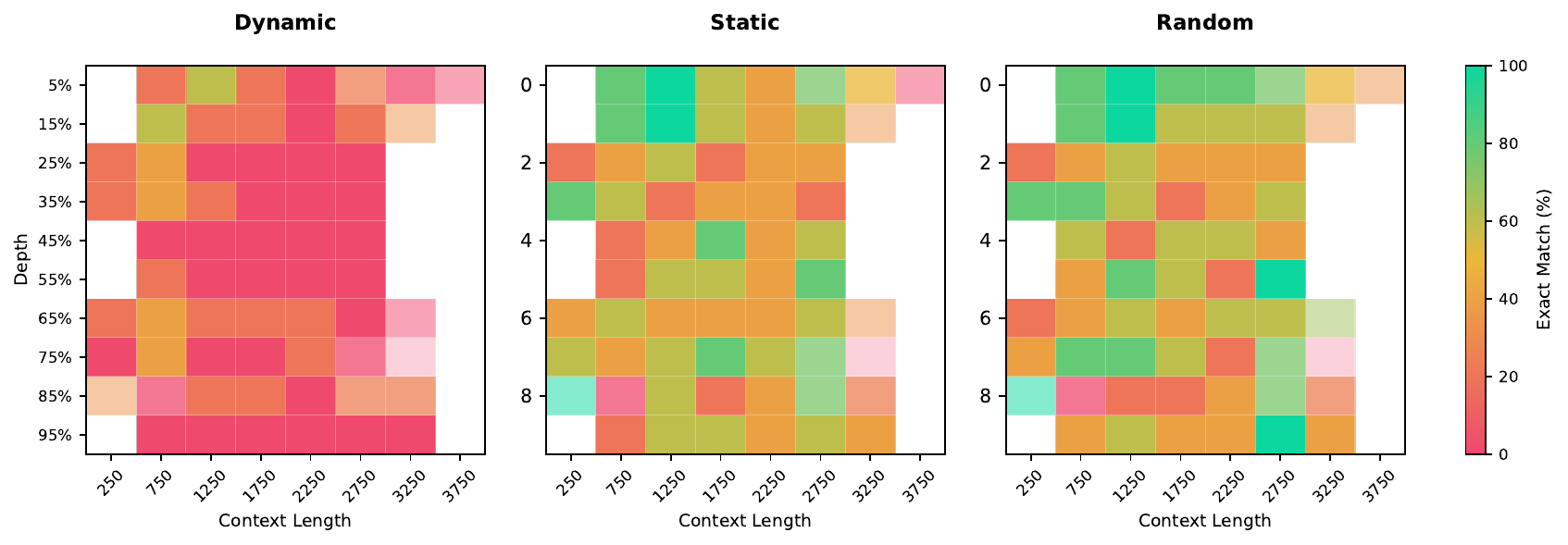}
    \caption{Head Ablation on HotpotQA test on llama3.1-8b. Using EM as the metric.}
    \label{fig:masking_hotpotqa_rouge}
\end{figure*}

\begin{figure*}
    \centering

    \begin{subfigure}{\textwidth}
        \centering
        \includegraphics[width=0.9\textwidth]{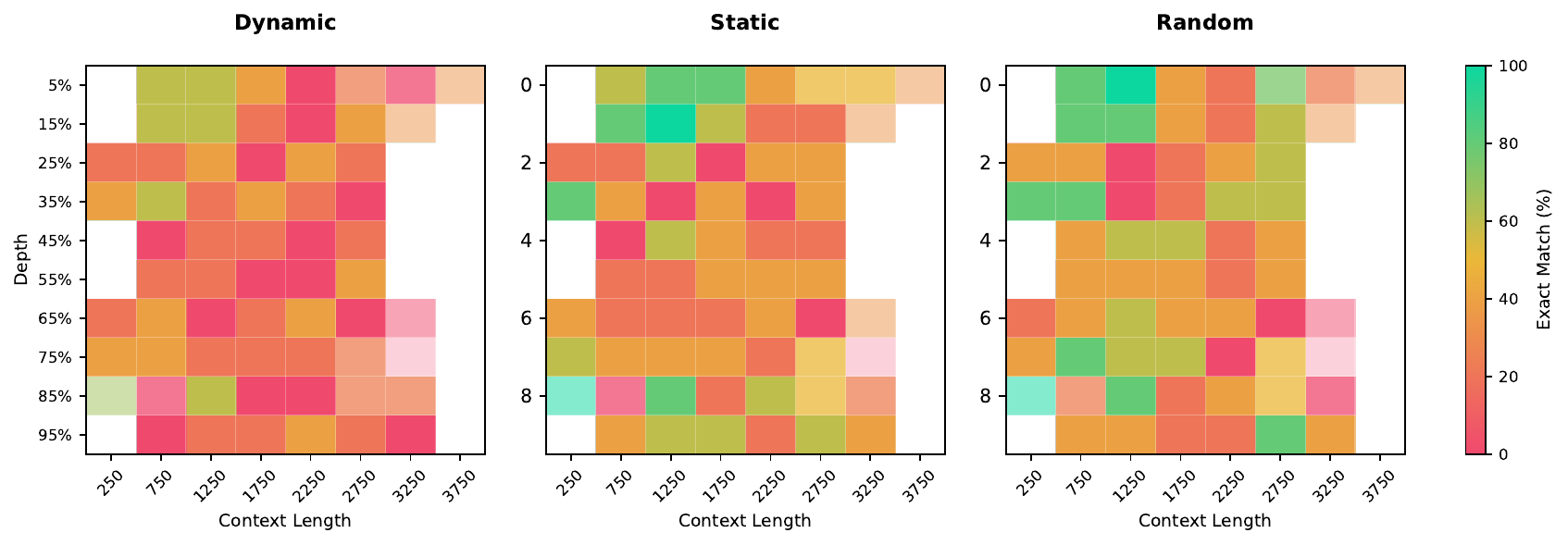}
        \caption{EM}
    \end{subfigure}
    
    \begin{subfigure}{\textwidth}
        \centering
        \includegraphics[width=0.9\textwidth]{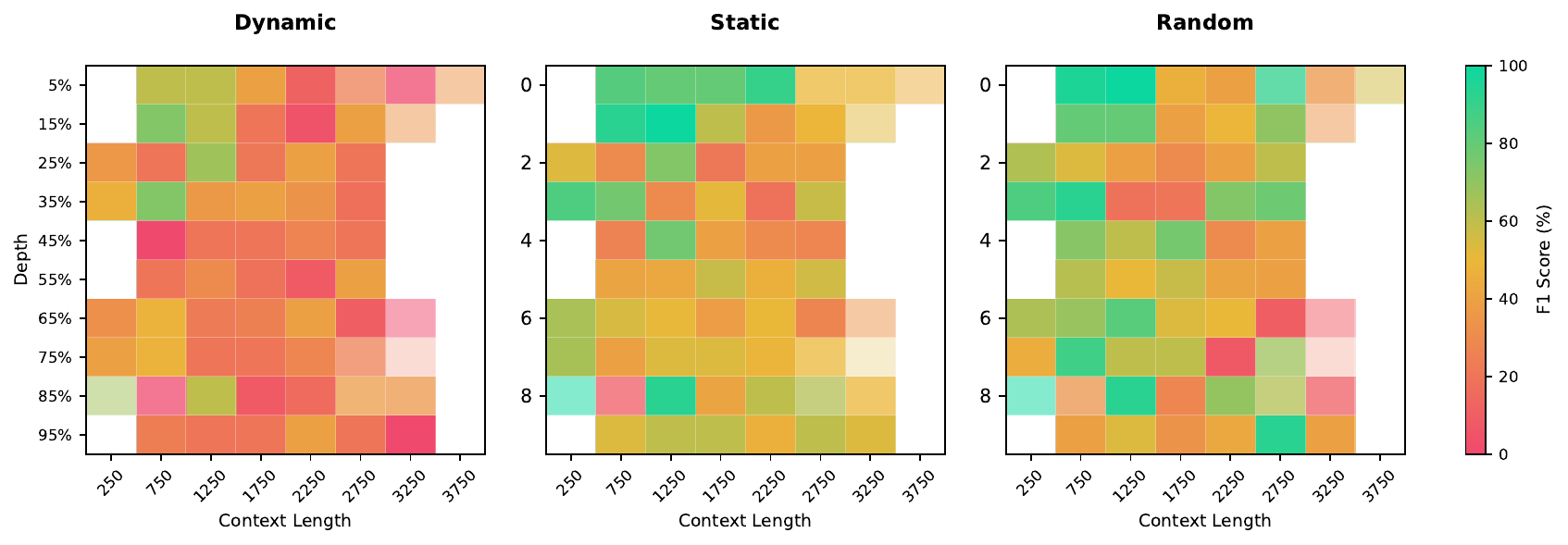}
        \caption{F1}
    \end{subfigure}

    \caption{Head Ablation on HotpotQA test on llama3.2-3b.}
    \label{fig:ap_masking_hotpotqa_llama3.2-3b}
\end{figure*}

\begin{figure*}
    \centering

    \begin{subfigure}{\textwidth}
        \centering
        \includegraphics[width=0.9\textwidth]{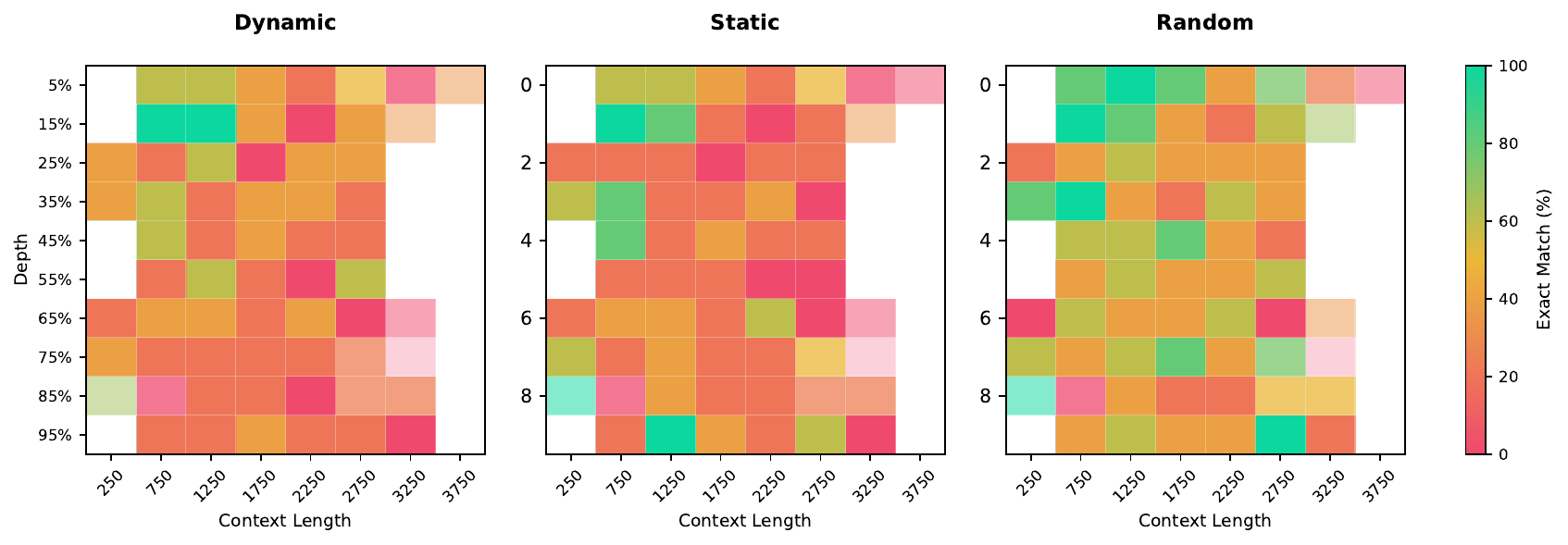}
        \caption{EM}
    \end{subfigure}
    
    \begin{subfigure}{\textwidth}
        \centering
        \includegraphics[width=0.9\textwidth]{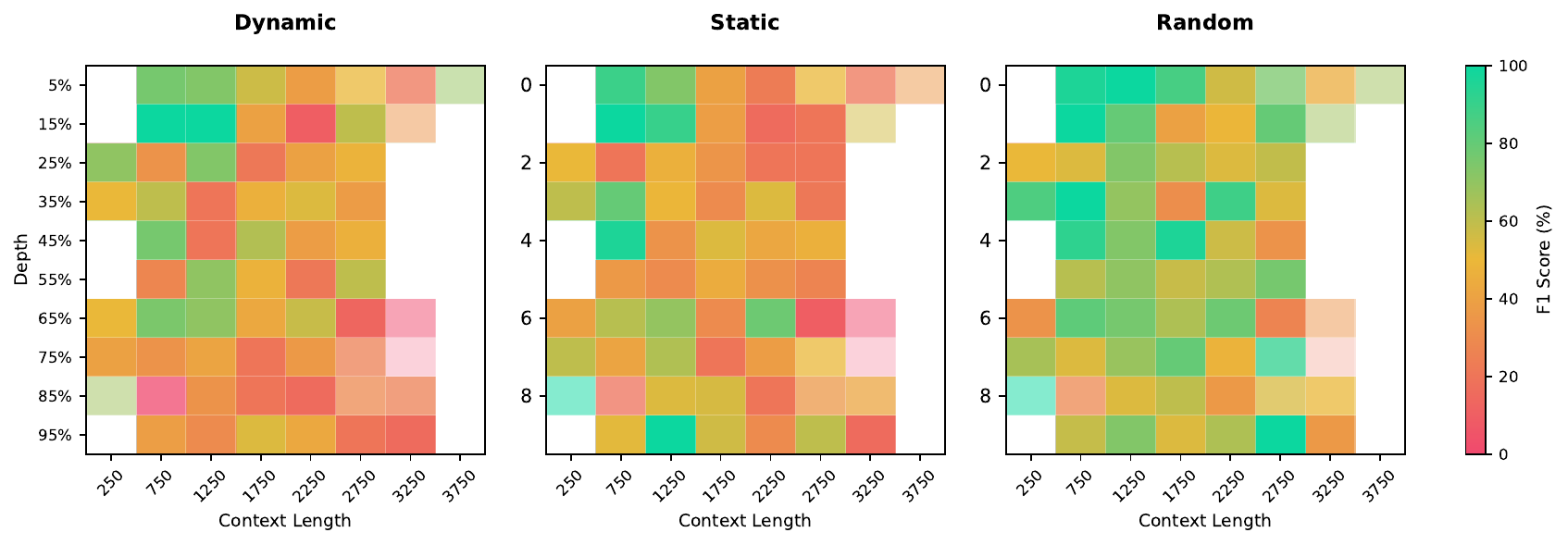}
        \caption{F1}
    \end{subfigure}

    \caption{Head Ablation on HotpotQA test on qwen3-8b.}
    \label{fig:ap_masking_hotpotqa_qwen3-8b}
\end{figure*}

See Figure~\ref{fig:dynamic_masking_rouge}, Figure~\ref{fig:ap_masking_llama3.2-3b}, Figure~\ref{fig:ap_masking_qwen3-8b}, Figure~\ref{fig:masking_hotpotqa_rouge}, Figure~\ref{fig:ap_masking_hotpotqa_llama3.2-3b}, Figure~\ref{fig:ap_masking_hotpotqa_qwen3-8b}.

\subsection{Different Numbers of Heads Ablation Results}\label{ap:niah_irreplaceability_result}

\begin{figure}
    \centering
    \includegraphics[width=0.9\linewidth]{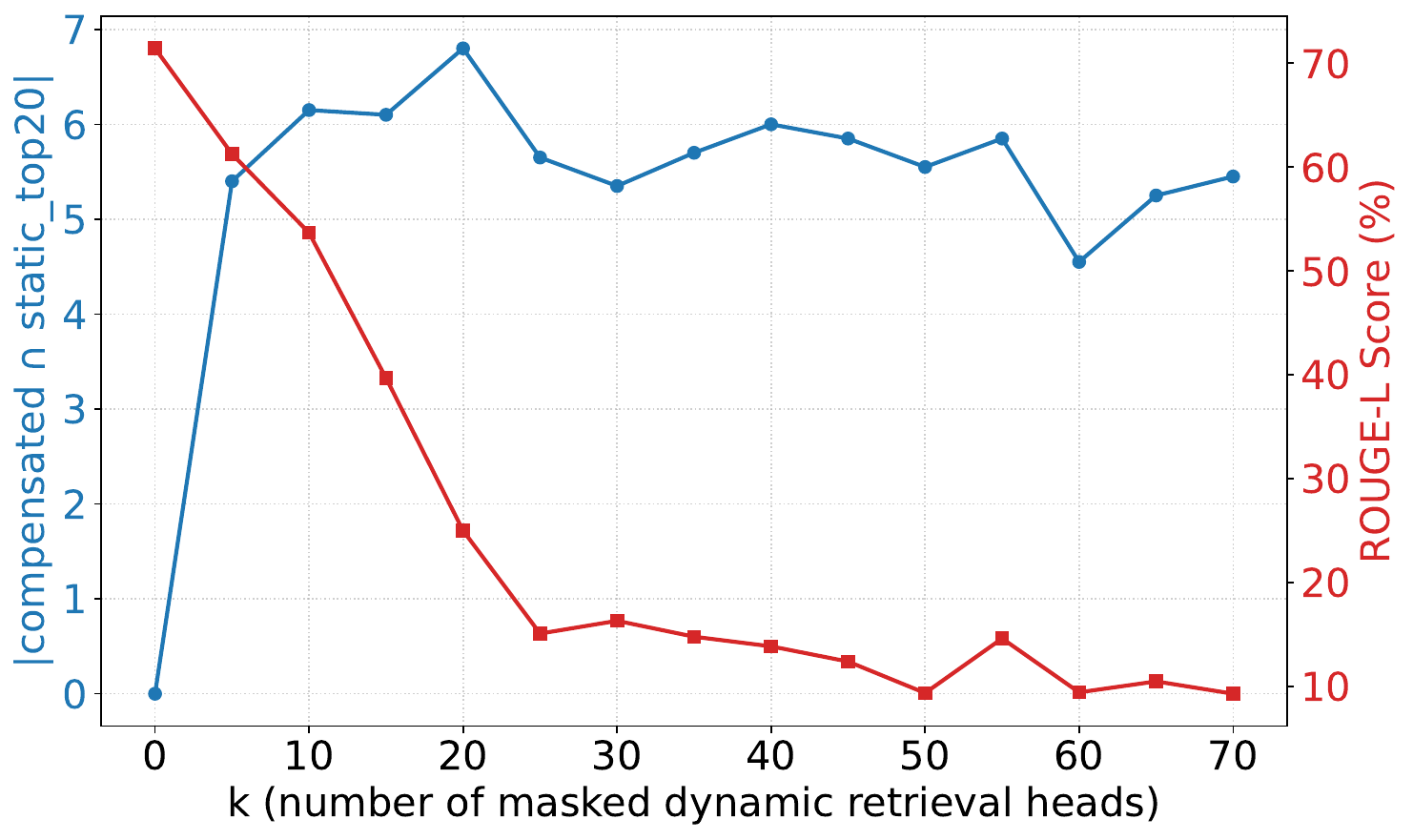}
    \caption{Different Numbers of Head Ablation on NIAH test on llama3.1-8b. Using ROUGE-L as the metric.}
    \label{fig:irr_em}
\end{figure}

\begin{figure}
    \centering

    \begin{subfigure}{\linewidth}
        \centering
        \includegraphics[width=0.9\linewidth]{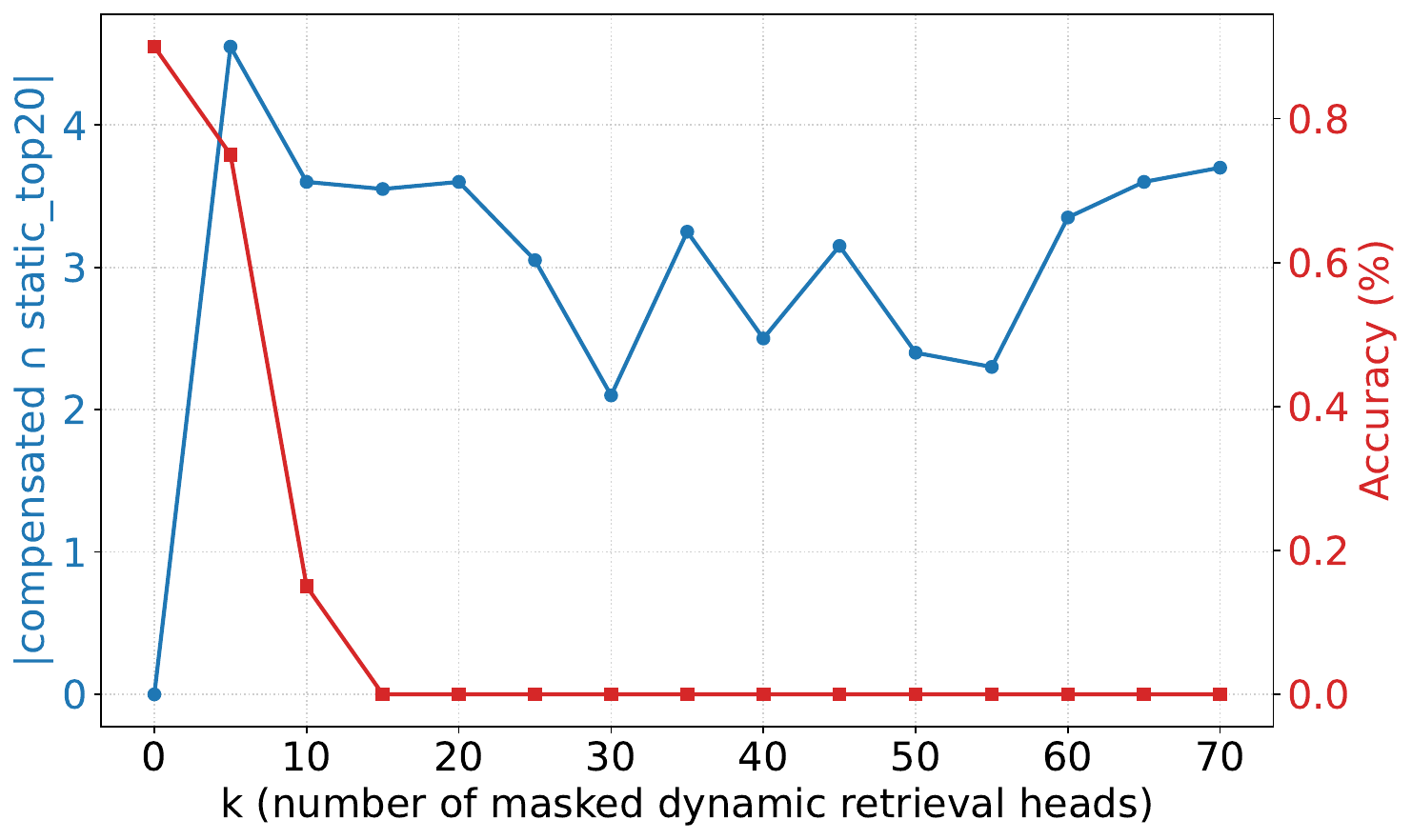}
        \caption{Accuracy}
    \end{subfigure}
    
    \vspace{5mm} 

    \begin{subfigure}{\linewidth}
        \centering
        \includegraphics[width=0.9\linewidth]{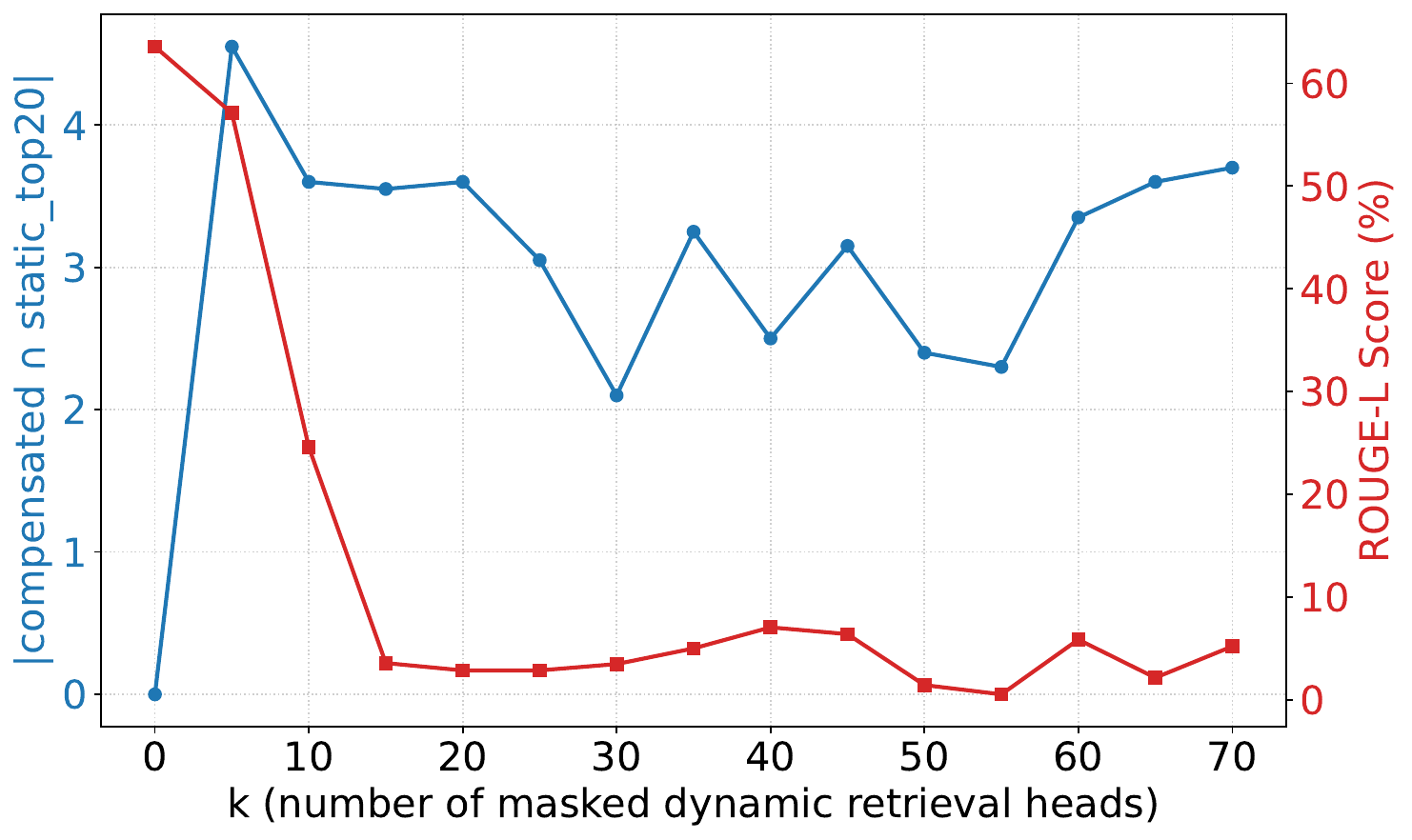}
        \caption{ROUGE-L}
    \end{subfigure}

    \caption{Different Numbers of Head Ablation on NIAH test on llama3.2-3b.}
    \label{fig:ap_irr_llama3.2-3b}
\end{figure}

\begin{figure}
    \centering

    \begin{subfigure}{\linewidth}
        \centering
        \includegraphics[width=0.9\linewidth]{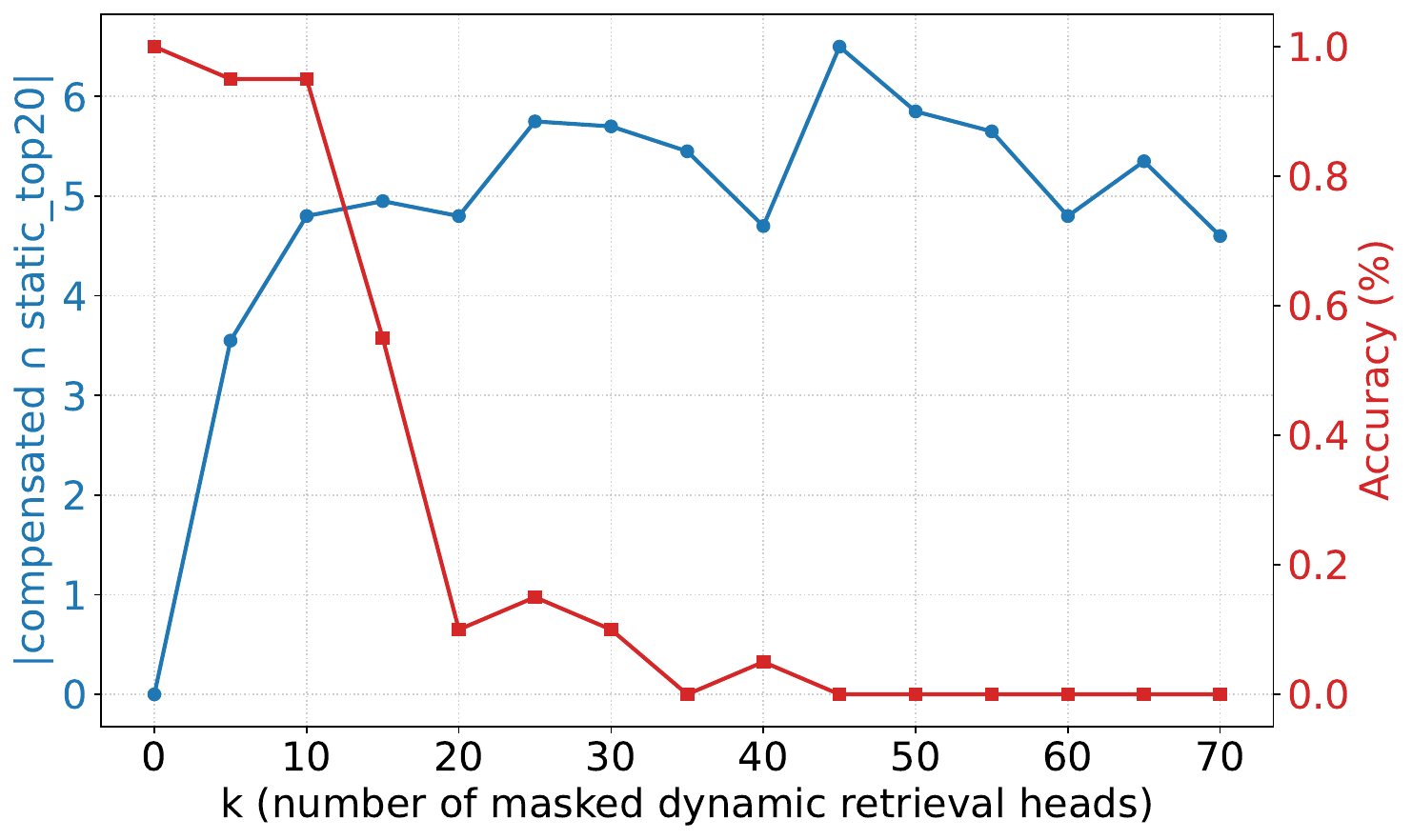}
        \caption{Accuracy}
    \end{subfigure}
    
    \vspace{5mm} 

    \begin{subfigure}{\linewidth}
        \centering
        \includegraphics[width=0.9\linewidth]{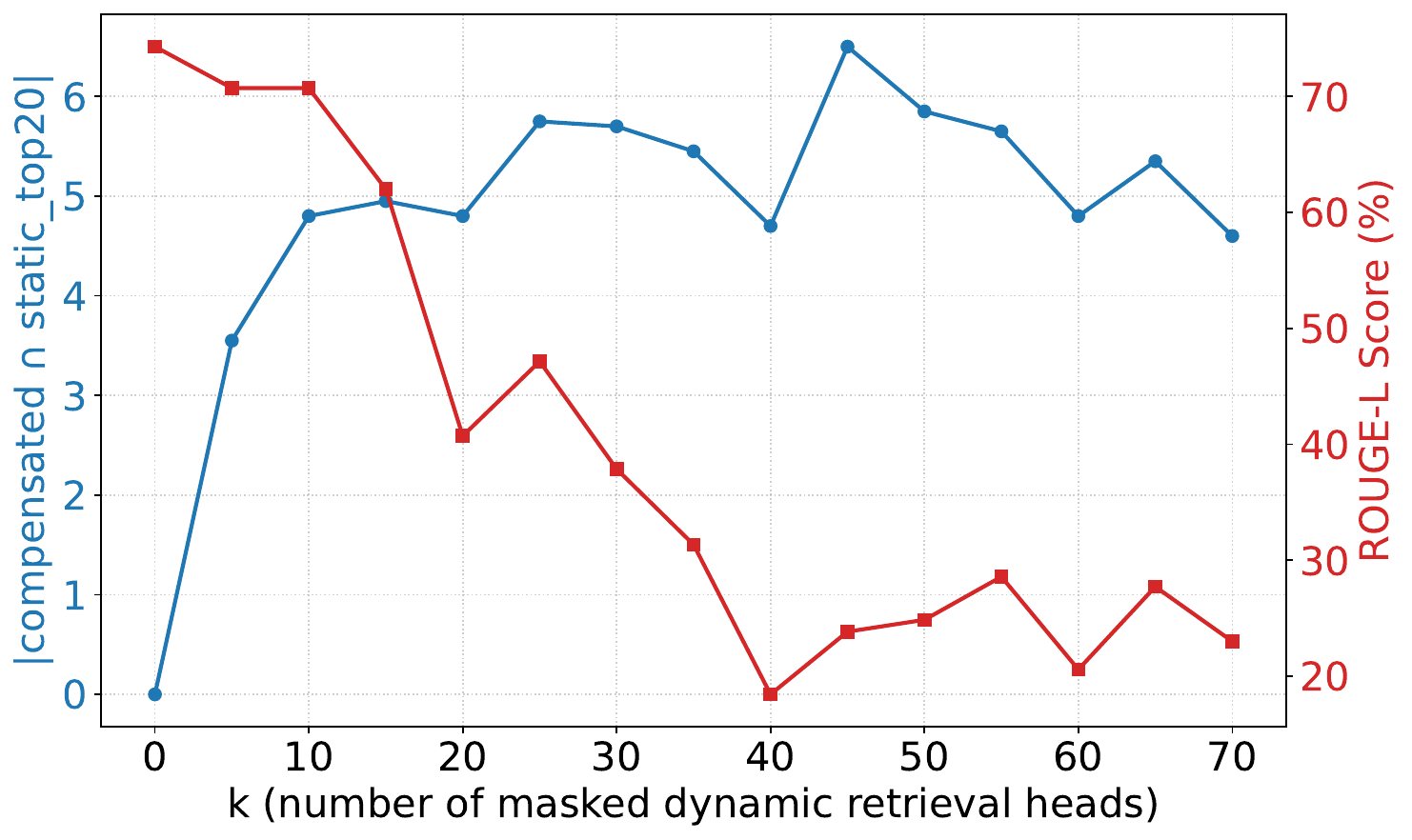}
        \caption{ROUGE-L}
    \end{subfigure}

    \caption{Different Numbers of Head Ablation on NIAH test on qwen3-8b.}
    \label{fig:ap_irr_qwen3-8b}
\end{figure}

\begin{figure}
    \centering
    \includegraphics[width=0.9\linewidth]{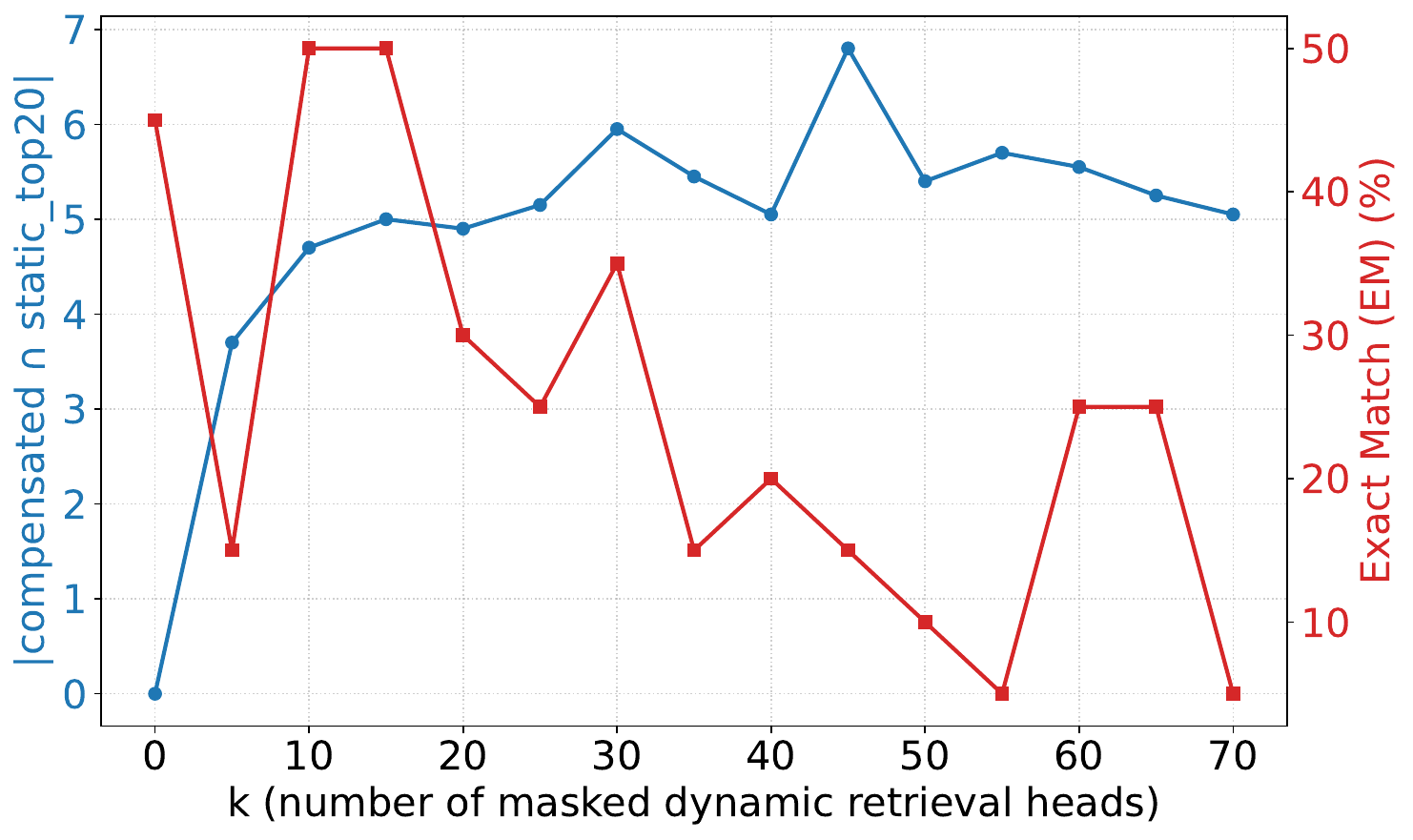}
    \caption{Different Numbers of Head Ablation on NIAH test on llama3.1-8b. Using EM as the metric.}
    \label{fig:irr_hotpotqa_em}
\end{figure}

\begin{figure}
    \centering

    \begin{subfigure}{\linewidth}
        \centering
        \includegraphics[width=0.9\linewidth]{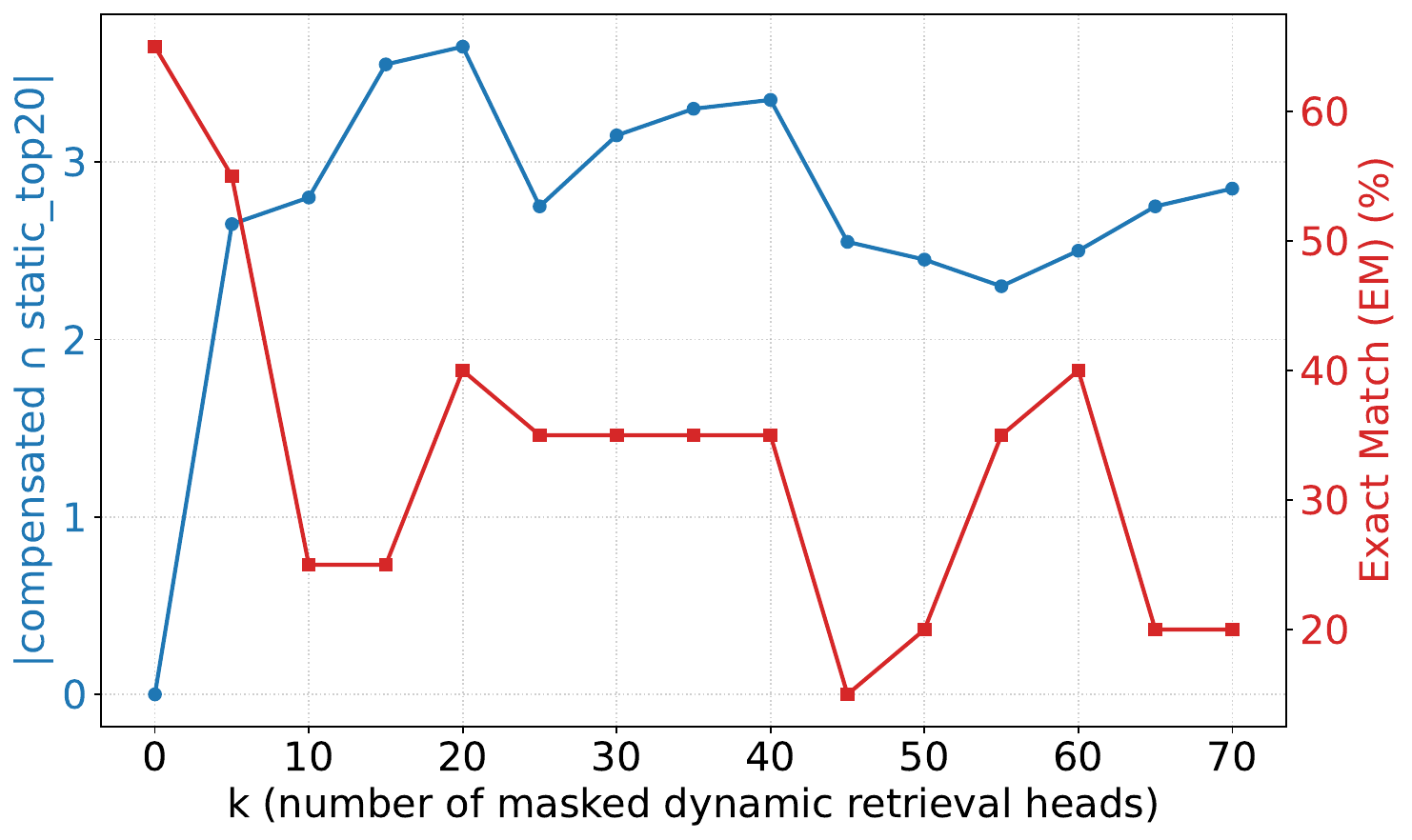}
        \caption{EM}
    \end{subfigure}
    
    \vspace{5mm} 

    \begin{subfigure}{\linewidth}
        \centering
        \includegraphics[width=0.9\linewidth]{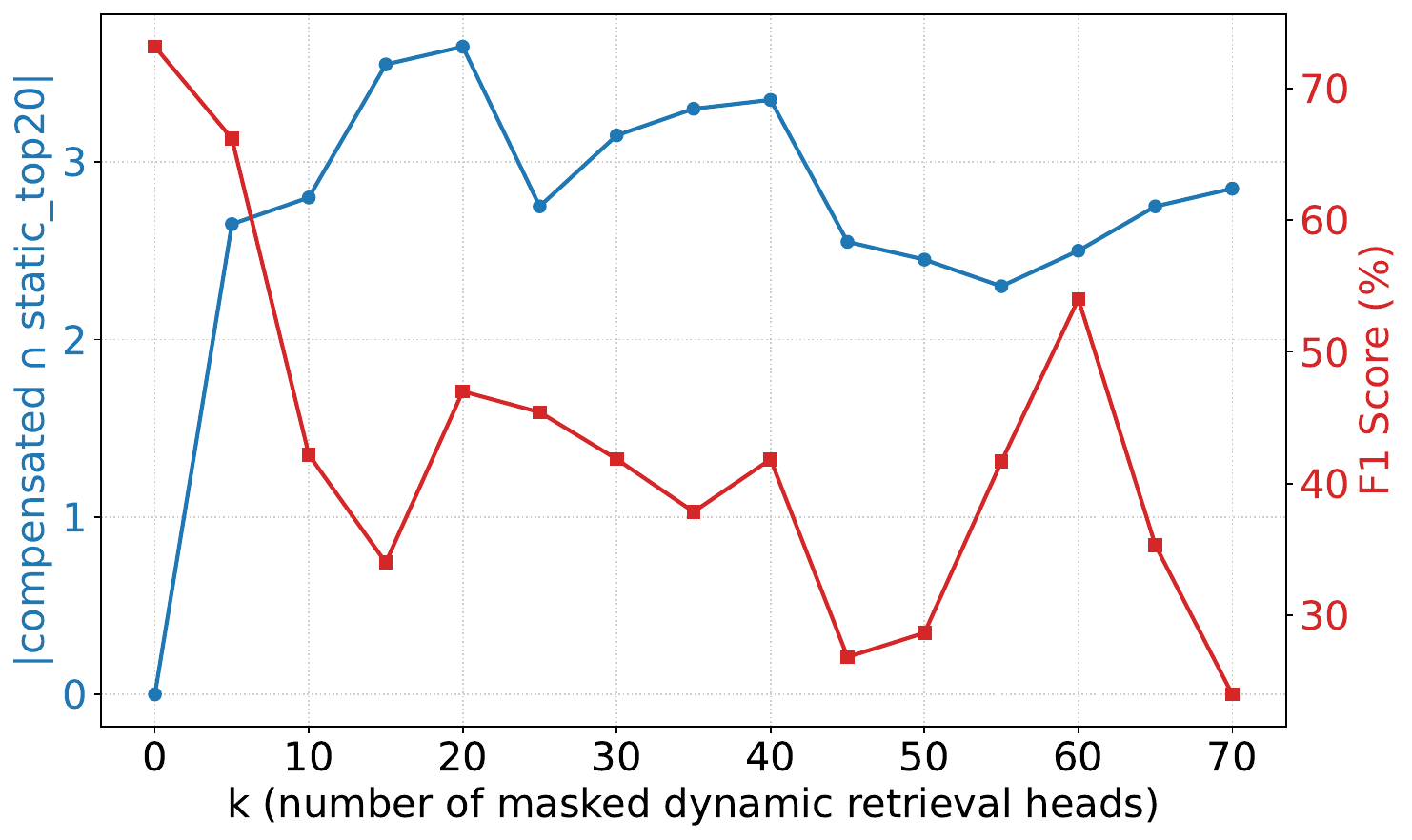}
        \caption{F1}
    \end{subfigure}

    \caption{Different Numbers of Head Ablation on HotpotQA test on llama3.2-3b.}
    \label{fig:ap_irr_hotpotqa_llama3.2-3b}
\end{figure}

\begin{figure}
    \centering

    \begin{subfigure}{\linewidth}
        \centering
        \includegraphics[width=0.9\linewidth]{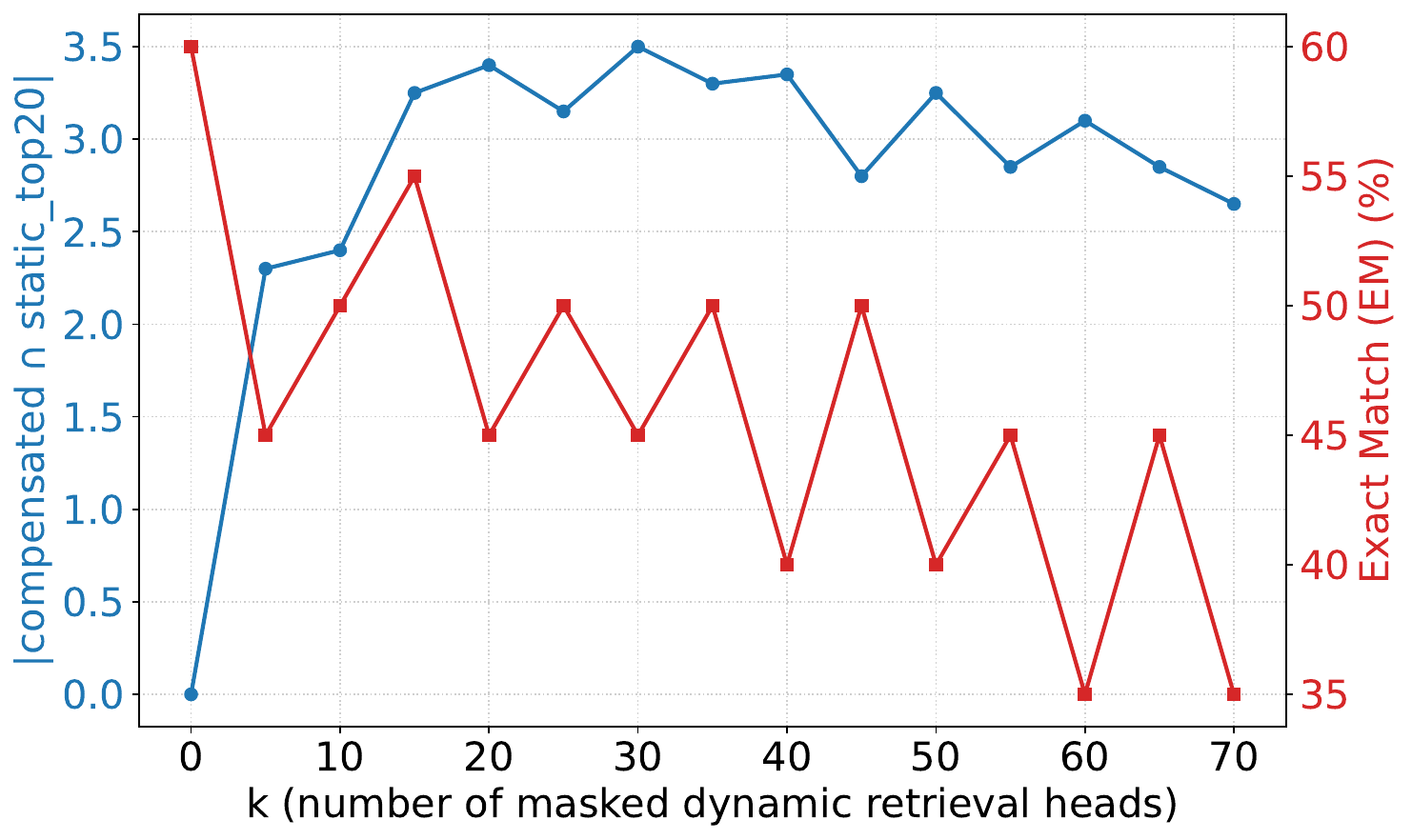}
        \caption{EM}
    \end{subfigure}
    
    \vspace{5mm} 

    \begin{subfigure}{\linewidth}
        \centering
        \includegraphics[width=0.9\linewidth]{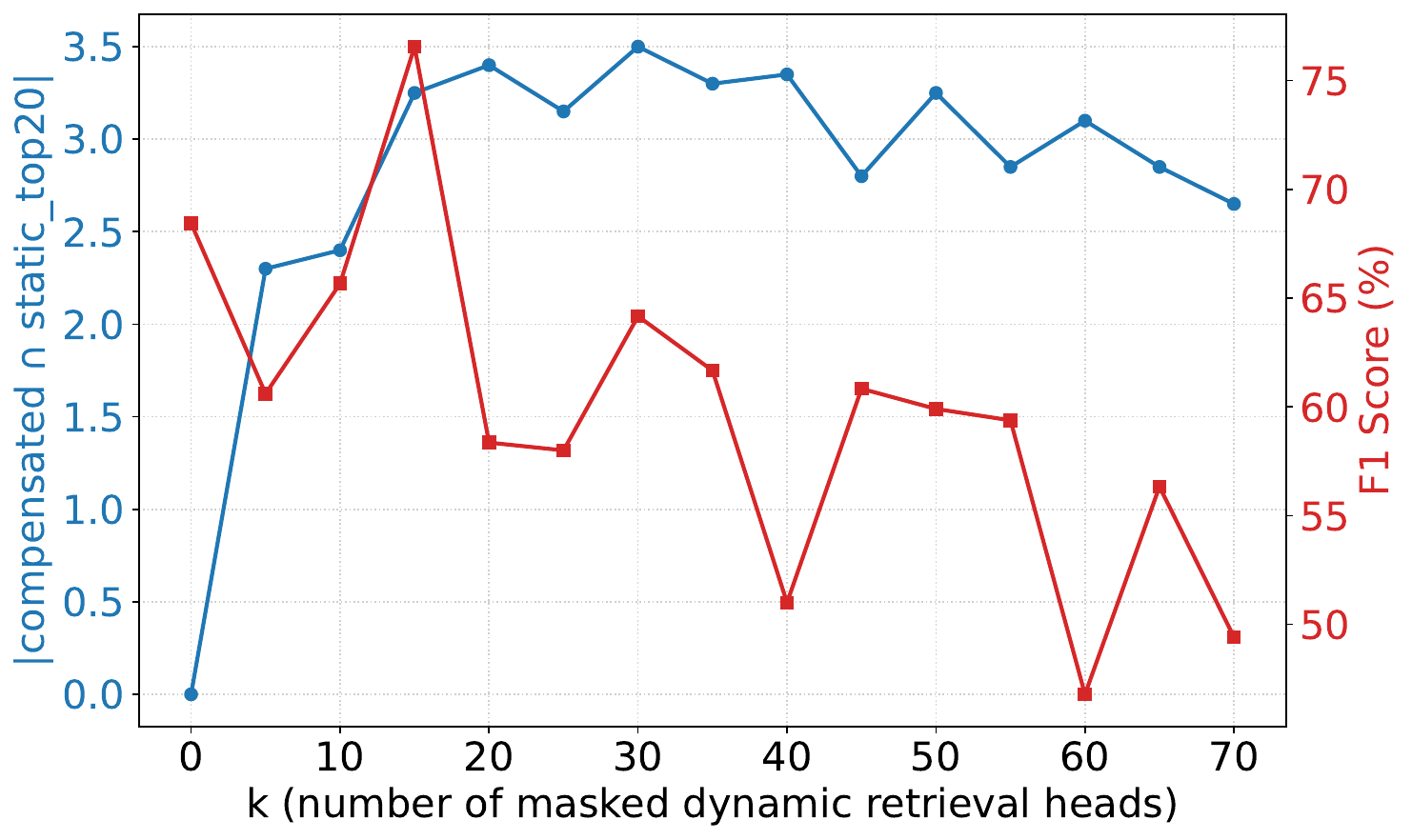}
        \caption{F1}
    \end{subfigure}

    \caption{Different Numbers of Head Ablation on HotpotQA test on qwen3-8b.}
    \label{fig:ap_irr_hotpotqa_qwen3-8b}
\end{figure}

See Figure~\ref{fig:irr_em}, Figure~\ref{fig:ap_irr_llama3.2-3b}, Figure~\ref{fig:ap_irr_qwen3-8b}, Figure~\ref{fig:irr_hotpotqa_em}, Figure~\ref{fig:ap_irr_hotpotqa_llama3.2-3b}, Figure~\ref{fig:ap_irr_hotpotqa_qwen3-8b}.

\section{Algorithms for the Dynamic RAG Method in Section~\ref{sec:experiment}}\label{ap:drag_algorithms}

See Algorithm~\ref{alg:main_loop}, Algorithm~\ref{alg:retrieve}.

\begin{algorithm*}[t]
\caption{Dynamic RAG with In-Context Retrieval}
\label{alg:main_loop}
\begin{algorithmic}[1]
\Require Context $\mathcal{C}$, Question $\mathcal{Q}$, Model $\mathcal{M}$
\State Generated Text $\mathcal{G} \gets \text{`'}$
\State Visible Mask $\mathcal{V} \gets \text{MaskAll}(\mathcal{C})$ \Comment{Initially mask context}

\While{not finished}
    \State Input $\mathcal{I} \gets \text{Concat}(\mathcal{C}, \mathcal{Q}, \mathcal{G})$
    \State Draft $\mathcal{D}$, Attentions $\mathcal{A} \gets \mathcal{M}.\text{GenerateDraft}(\mathcal{I}, \text{mask}=\mathcal{V})$ 
    
    \State is\_hallucination, pos $\gets \text{RIND}(\mathcal{D})$
    
    \If{is\_hallucination}
        \State $\mathcal{G} \gets \text{Retract}(\mathcal{G}, \text{to sentence of } \text{pos})$
        \State $\mathcal{V} \gets \textbf{Retrieve}(\mathcal{C}, \mathcal{Q}, \mathcal{G})$ \Comment{See Alg.~\ref{alg:retrieve}}
        \State Input $\mathcal{I} \gets \text{Concat}(\mathcal{C}, \mathcal{Q}, \mathcal{G})$
        \State Rewritten $\gets \mathcal{M}.\text{Generate}(\mathcal{I}, \text{mask}=\mathcal{V})$
        \State $\mathcal{G} \gets \text{Append}(\mathcal{G}, \text{Rewritten})$
    \Else
        \State $\mathcal{G} \gets \text{Append}(\mathcal{G}, \mathcal{D})$
        \State $\mathcal{V} \gets \text{MaskAll}(\mathcal{C})$ \Comment{Re-mask for next draft}
    \EndIf
\EndWhile
\State \Return $\mathcal{G}$
\end{algorithmic}
\end{algorithm*}

\begin{algorithm*}[t]
\caption{In-Context Retrieval via Attention Heads}
\label{alg:retrieve}
\begin{algorithmic}[1]
\Function{Retrieve}{$\mathcal{C}, \mathcal{Q}, \mathcal{G}$}
    \State UnMaskAll($\mathcal{C}$)
    \State Active Heads $\mathcal{H}_{dyn} \gets \text{IdentifyHeads}(\mathcal{C}, \mathcal{Q}, \mathcal{G})$
    \State MaskAll($\mathcal{C}$)
    \State Avg Scores $\mathbf{s} \gets \text{AverageAttention}(\mathcal{H}_{dyn})$
    \State Top-k Indices $\mathcal{K} \gets \text{TopKIndices}(\mathbf{s}, k)$
    
    \State Index Clusters $\mathcal{C}_{idx} \gets \text{ClusterIndices}(\mathcal{K})$
    
    \State Expanded Windows $\mathcal{W} \gets \emptyset$
    \For{each cluster $c \in \mathcal{C}_{idx}$}
        \State Representative Index $i_{rep} \gets \text{GetRepresentative}(c)$
        \State $\mathcal{W} \gets \mathcal{W} \cup \text{ExpandWindow}(i_{rep}, \text{size})$
    \EndFor
    
    \State Final Windows $\mathcal{W}_{final} \gets \text{MergeOverlapping}(\mathcal{W})$
    
    \State Visible Mask $\mathcal{V} \gets \text{CreateMaskFromWindows}(\mathcal{W}_{final})$
    \State \Return $\mathcal{V}$
\EndFunction
\end{algorithmic}
\end{algorithm*}

\end{document}